\documentclass[runningheads]{llncs}

\usepackage{eccv}

\usepackage{eccvabbrv}

\usepackage{graphicx}
\usepackage{booktabs}

\usepackage{pifont}%
\newcommand{\cmark}{{\color{ForestGreen} \ding{51}}}%
\newcommand{\xmark}{{\color{red} \ding{55}}}%

\newcommand{\rom}[1]{\uppercase\expandafter{\romannumeral #1\relax}}

\usepackage{xcolor}

\usepackage{calc}

\usepackage[accsupp]{axessibility}  %
\usepackage{colortbl}
\usepackage{tabulary}
\usepackage{tabularx}
\usepackage{etoolbox}
\usepackage{multirow}
\usepackage{cuted}
\usepackage[percentage]{overpic}
\usepackage{booktabs}
\usepackage{pgfplots}
\pgfplotsset{compat=newest}%
\usepackage{algorithm}
\usepackage{algpseudocode}
\usepackage{amsmath}
\usepackage{bbding}

\usepackage{stackrel}

\usepackage{tikz}
\usetikzlibrary{calc}
\usetikzlibrary{decorations.pathmorphing}
\usetikzlibrary{patterns,arrows,automata,positioning,decorations,shapes,calc}

\tikzstyle{normalvertex}=[circle,fill=white,draw=black]
\tikzstyle{emptyvertex}=[draw,circle,minimum size=7pt,inner sep=0pt]
\tikzstyle{tinyvertex}=[draw,circle,minimum size=3pt,inner sep=0pt]
\tikzstyle{thickedge}=[draw,gray!60,line width=1.6pt,-]

\usepackage{pgfplots}
\usepgfplotslibrary{colorbrewer}

\pgfdeclarelayer{background}
\pgfsetlayers{background,main}

\usetikzlibrary{shapes}
\usetikzlibrary{arrows.meta,backgrounds,automata, chains}
\usetikzlibrary{positioning,shapes,matrix,calc}
\tikzstyle{vertex}=[circle, draw, fill=gray!80!white,thick,scale=1.2]
\tikzstyle{edge}=[draw=black, thick,-]

\usepackage{bm}

\definecolor{purple}{RGB}{147,7,204}
\definecolor{blue}{RGB}{10,153,201}
\definecolor{orange}{RGB}{254,128,41}
\definecolor{gray}{RGB}{239,240,241}
\definecolor{pink}{RGB}{254,15,127}
\definecolor{green}{RGB}{140,211,89}

\definecolor{color1}{RGB}{254,15,127}
\definecolor{color2}{RGB}{10,153,201}
\definecolor{color3}{RGB}{194,145,162}
\definecolor{color4}{RGB}{254,128,41}
\definecolor{color5}{RGB}{254,191,185}
\definecolor{color6}{RGB}{110,231,169}
\definecolor{color7}{RGB}{245,221,66}

\newcommand{\M}{\mathcal{M}}
\newcommand{\N}{\mathcal{N}}

\newcommand{\yes}{\color{green}{\CheckmarkBold}}
\newcommand{\no}{\color{red}{\XSolidBrush}}

\def\pathOurs{figs/ours/}
\def\srcEnd{_M}
\def\trgtEnd{_N}

\usepackage[pagebackref,breaklinks,colorlinks,citecolor=eccvblue]{hyperref}

\usepackage{orcidlink}

\begin{document}
\setkeys{Gin}{keepaspectratio}
\title{Synchronous Diffusion for Unsupervised Smooth Non-Rigid 3D Shape Matching} 

\titlerunning{Synchronous Diffusion for Unsupervised 3D Shape Matching}

\author{Dongliang Cao\inst{1}\orcidlink{0000-0002-6505-6465} \and
Zorah Lähner\inst{1,2}\orcidlink{0000-0003-0599-094X} \and
Florian Bernard\inst{1}\orcidlink{0009-0008-1137-0003}}

\authorrunning{Dongliang Cao et al.}

\institute{\textsuperscript{1}University of Bonn \quad \textsuperscript{2}Lamarr Institute}

\maketitle

\begin{abstract}
  Most recent unsupervised non-rigid 3D shape matching methods are based on the functional map framework due to its efficiency and superior performance. Nevertheless, respective methods struggle to obtain spatially smooth pointwise correspondences  due to the lack of proper regularisation. In this work, inspired by the success of message passing on graphs, we propose a \emph{synchronous diffusion process} which we use as regularisation to achieve smoothness in non-rigid 3D shape matching problems. The intuition of synchronous diffusion is that diffusing the same input function on two different shapes results in consistent outputs. Using different challenging datasets, we demonstrate that our novel regularisation can substantially improve the state-of-the-art in shape matching, especially in the presence of topological noise.
\end{abstract}

\begin{figure}[bht!]
    \centering
    \includegraphics[width=\textwidth]{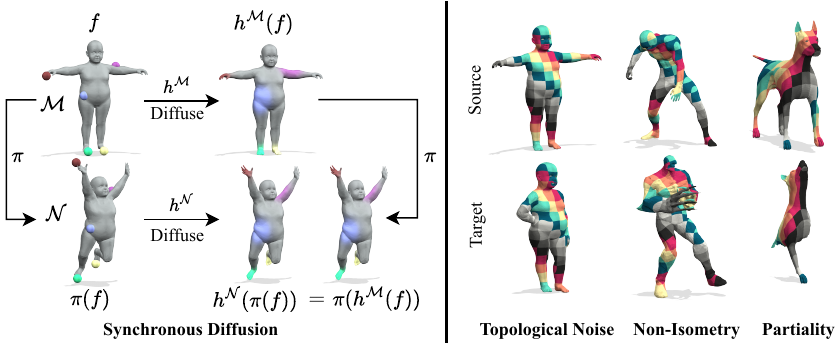}
    \caption{\textbf{Left:} We present an unsupervised regularisation based on \emph{synchronous diffusion} for spatially smooth non-rigid 3D shape matching. This is based on the motivation that diffusing a function $f$ on shape $\mathcal{M}$ should lead to comparable results when diffusing $\pi(f)$ (i.e.~$f$ transferred to shape $\mathcal{N}$ using the map $\pi$) on shape $\mathcal{N}$. Coloured points are used to illustrate a (subset of) random function values.
    \textbf{Right:} Our approach can be applied in a broad range of challenging scenarios, including topological noise, non-isometry and partiality.}
    \vspace{-6mm}
    \label{fig:teaser}
\end{figure}

\section{Introduction}
\label{sec:intro}
Finding correspondences between non-rigid 3D shapes is a fundamental problem in computer vision and computer graphics with a diverse range of applications, including texture transfer~\cite{dinh2005texture}, pose transfer~\cite{song20213d,song2023unsupervised} and statistical shape analysis~\cite{loper2015smpl,li2017flame,egger20203d}, to name just a few. Even though non-rigid 3D shape matching is a long-standing problem and has been studied for decades~\cite{deng2022survey,sahilliouglu2020recent}, finding accurate pointwise correspondence remains challenging, especially in the presence of topological noise and partiality, which are common in real-world 3D scan data. 

Inspiringly, recent deep functional map methods witnessed great improvements for non-rigid 3D shape matching including near-isometric shape matching~\cite{attaiki2023understanding}, non-isometric shape matching~\cite{donati2022deep,li2022learning}, multi-shape matching~\cite{cao2022unsupervised,eisenberger2023g,sun2023spatially} as well as partial shape matching~\cite{attaiki2021dpfm}. Moreover, a recent deep functional map framework~\cite{cao2023unsupervised,cao2024revisiting} achieves state-of-the-art matching performance in diverse settings. Despite the success of deep functional map methods,  they map shapes in the low-frequency spectral domain and thus often lead to locally inconsistent pointwise correspondences. To address this limitation, many works~\cite{eisenberger2021neuromorph,jiang2023non,attaiki2023shape,cao2024spectral} attempt to additionally register shapes in the spatial domain to ensure spatial smoothness of the obtained pointwise correspondences. Nevertheless, these methods require that the given shapes are already rigidly aligned and they are sensitive to topological noise and not directly applicable for partial shapes. 

In contrast, our objective is to design a universal regularisation that encourages spatially smooth matching and can be easily integrated into existing non-rigid 3D shape matching methods~\cite{attaiki2023understanding,sun2023spatially,cao2023unsupervised}. Prior works that ensure smoothness~\cite{ren2018continuous,ezuz2019reversible} typically require computationally expensive iterative optimisation and thus cannot be integrated straightforwardly into learning frameworks. To this end, inspired by message passing mechanisms used in the graph isomorphism test~\cite{weisfeiler1968reduction} and for deep graph matching~\cite{fey2019deep}, we propose \emph{synchronous diffusion} as smoothness regularisation for non-rigid 3D shape matching. In contrast to message passing mechanisms that aggregate information based on local connectivity (e.g.\ one-hop neighbourhood), the diffusion process~\cite{kondor2002diffusion} utilises smooth communication~\cite{gasteiger2019diffusion,defferrard2016convolutional}, which is more robust to discretisation~\cite{vallet2008spectral,zhang2010spectral,sharp2020diffusionnet} and thus more suitable for 3D shape analysis.

In the following we formally define synchronous diffusion:
\begin{definition}[Synchronous diffusion]\label{def:syncdiff}
Let $\mathcal{M}$ and $\mathcal{N}$ be two domains (e.g.~3D shapes) and let $\pi$ be an invertible mapping from functions defined on $\mathcal{M}$ to $\mathcal{N}$. We say that the diffusion processes $h^\mathcal{M}$ (diffusion on $\mathcal{M}$) and $h^\mathcal{N}$   (diffusion on $\mathcal{N}$) are \emph{synchronous} (w.r.t.~$\pi$) if for any function $f$ defined on $\mathcal{M}$ the following property holds: $\pi(h^\mathcal{M}(f)) = h^\mathcal{N}(\pi(f))$.
\end{definition}
Intuitively, the idea of \emph{synchronous diffusion} is to apply a diffusion process to the `same' function on different domains (3D shapes in our case), see~\cref{fig:teaser} left for an illustration. Our key observation is that  synchronous diffusion favours the mapping to be globally continuous and locally smooth. Therefore, we propose a novel regularisation based on synchronous diffusion to encourage spatially smooth pointwise correspondences. 

\cref{tab:comparison} compares our proposed method with the most relevant existing approaches. In contrast to prior works, our method is the first approach that is specifically designed for unsupervised smooth 3D shape matching with a unique combination of desirable properties. In \cref{sec:theoretical} we analyse our proposed regularisation and build connections with well-established heat kernel matching~\cite{vestner2017efficient} and  the Weisfeiler-Leman graph isomorphism test~\cite{weisfeiler1968reduction}. In our experiments, we demonstrate that our regularisation can be integrated into the state-of-the-art deep functional map method~\cite{cao2023unsupervised} and substantially improve its performance. We summarise our main contributions as follows:

\begin{itemize}
    \item For the first time we propose a simple yet efficient regularisation for unsupervised 3D shape matching that encourages spatially smooth pointwise correspondences under a broad range of challenging scenarios.
    \item To this end, we introduce a \emph{synchronous diffusion} process that penalises local mismatches without requiring ground truth correspondences.
    \item We demonstrate that our regularisation substantially improves the state-of-the-art deep functional map method in various challenging problem settings.
\end{itemize}

\begin{table}[t!]
\scriptsize
\setlength{\tabcolsep}{3.5pt}
    \centering
    \caption{Comparison to related existing methods. Our method is the first approach that is based on multiscale \emph{synchronous} heat diffusion, tailored specifically for unsupervised smooth 3D shape matching, while being more efficient than previous kernel matching methods based on quadratic assignment formulations.
    \vspace{-2mm}
    }
    \label{tab:comparison}
    \begin{tabular}{lccccc}
    \toprule
                                                        & Heat diffusion   & Efficient & Multiscale     & Smooth & Synchronous\\ \midrule
        Heat Kernel Signature \cite{bronstein2010scale} & \yes      & \yes   & \yes        & \no            & \no \\
        DiffusionNet \cite{sharp2020diffusionnet}       & \yes      & \yes  & \yes       & \no      & \no \\
        Dirichlet Energy \cite{ezuz2019reversible,magnet2022smooth} & \no      & \yes   & \no         & \yes    & \no \\
        Kernel Matching \cite{vestner2017efficient}     & \yes      & \no    & \yes        & \yes      & \no \\
        Unsup FMNet \cite{halimi2019unsupervised} & \no      & \no   & \no        & \yes      & \no \\
        Heat Kernel UDFMs \cite{aygun2020heatkernel}    & \yes      & \no   & \yes         & \yes       & \no \\
        Ours                                            & \yes      & \yes  & \yes        & \yes    &  \yes \\
    \bottomrule
    \end{tabular}
    \vspace{-4mm}
\end{table}

\section{Related Work}
\label{sec:related_work}
\subsection{Functional Map Methods for Non-Rigid 3D Shape Matching}
Non-rigid 3D shape matching is a well-studied problem in computer vision and graphics~\cite{van2011survey,tam2012registration}. Rather than providing an exhaustive literature survey, in the following we will focus on reviewing the most relevant methods that are based on the functional map framework~\cite{ovsjanikov2012functional}. Instead of directly finding pointwise correspondences, which is often formulated as expensive combinatorial optimisation problem~\cite{windheuser2011geometrically,bernard2020mina,holzschuh2020simulated,roetzer2022scalable,gao2023sigma}, the functional map framework finds correspondences in the functional domain. Here, the correspondence relationship can be encoded into a small matrix, namely the functional map, by using a series of truncated basis functions~\cite{ovsjanikov2012functional} (typically the first $k$ Laplacian eigenfunctions~\cite{levy2006laplace}). Due to its simplicity and efficiency, the functional map framework has been extended in numerous follow-up works, e.g.~in terms of improving the accuracy or robustness~\cite{eynard2016coupled,ren2019structured,melzi2019zoomout}, as well as extending its application to partial shape matching~\cite{rodola2017partial,litany2017fully}, non-isometric shape matching~\cite{nogneng2017informative,ren2018continuous,ren2021discrete}, multi-shape matching~\cite{huang2014functional,huang2020consistent,gao2021isometric} and matching with non-unique solutions~\cite{ren2020maptree}. Nevertheless, axiomatic functional map methods typically rely on handcrafted features~(e.g.\ HKS~\cite{bronstein2010scale}, WKS~\cite{aubry2011wave}). To further improve the matching performance, many learning-based methods propose to learn the features directly from the training data, either with supervision from ground-truth correspondences~\cite{litany2017deep,donati2020deep} or without supervision based on isometry assumptions~\cite{halimi2019unsupervised,roufosse2019unsupervised}. However, deep functional map methods often obtain local erroneous pointwise correspondences due to the use of low-frequency basis functions and the lack of proper regularisation. To address this limitation, we propose an efficient regularisation based on the diffusion process that encourages spatially smooth correspondences.
\vspace{-2mm}
\subsection{Message Passing for Graph Analysis}
The message passing mechanism is an important tool for graph analysis. Despite its simplicity (i.e.\ aggregate neighbouring information and update the node features), message passing mechanisms are widely used for graph analysis. Use cases range from traditional graph kernel analysis~\cite{shervashidze2009fast,rogers2010extended} for graph embedding, to recent graph neural networks (GNNs)~\cite{wu2020comprehensive,xu2018powerful} for graph representation learning~\cite{gilmer2017neural,hamilton2017inductive,morris2019weisfeiler}. In the context of the graph isomorphism problem (i.e.\ test whether two graphs have the same topological structure), the Weisfeiler-Leman test (WL test)~\cite{weisfeiler1968reduction} is a well-known heuristic test based on message passing and has recently gained popularity in the context of graph neural networks \cite{morris2021weisfeiler,franks2024weisfeiler}. The WL test iteratively checks for node label consistency between two graphs in each message passing step.  A more comprehensive review about message passing for graph analysis can be found in~\cite{morris2021weisfeiler}. The most relevant application of message passing to our work is deep graph matching. Recent deep graph matching methods~\cite{zanfir2018deep,fey2019deep,gao2021deep,tourani2023unsupervised} find node correspondences based on pairwise node feature similarities. To encourage neighbourhood consensus (i.e. neighbouring nodes are still neighbouring nodes after matching), DGMC~\cite{fey2019deep} proposes a method to iteratively improve the neighbourhood consensus in a subsequent second refinement stage via a differentiable validator (i.e.\ an additional GNN) for graph isomorphism based on the WL test heuristic. Similar to deep graph matching methods, many recent deep functional map methods~\cite{attaiki2023ncp,cao2023self,cao2023unsupervised} also obtain pointwise correspondences based on pairwise feature similarities. To this end, our work is inspired by the idea of using message passing for consensus graph matching in DGMC. However, instead of using the computationally expensive second refinement stage, we propose an efficient regularisation based on synchronous diffusion that is seamlessly integrated into the current deep shape matching framework~\cite{cao2023unsupervised}, which in turn significantly accelerates both training and inference, while having better scalability to high-resolution shapes.   

\subsection{Diffusion  for 3D Shape Analysis}
The diffusion process, which originates from the analysis of heat transfer, enables various applications in the context of 3D shape analysis, including shape segmentation~\cite{rustamov2007laplace}, shape matching~\cite{ovsjanikov2010one,ovsjanikov2012functional}, multi-resolution shape representation~\cite{levy2006laplace}, and geometric deep learning~\cite{sharp2020diffusionnet}, among many other applications (see~\cite{vallet2008spectral} for an overview). In the context of shape matching, heat kernel matching~\cite{vestner2017efficient} is highly relevant to our work. In this work, a feature-based linear assignment problem (LAP) and a kernel-based quadratic assignment problem (QAP) are solved together in an iterative way. Unlike previous work that uses geodesic distance kernels~\cite{vestner2017product}, heat kernels (i.e.\ the solutions to the heat diffusion equations) are chosen in this method due to their computational efficiency, their bias towards \emph{local} correspondences, and because they are more robust to topological noise~\cite{vestner2017efficient}. The main difference between our regularisation and heat kernel matching is that heat kernel matching measures the distances in the kernel space, while our regularisation penalises unsmooth correspondences in the feature space, which is more efficient~\cite{fey2019deep}. Moreover, the heat kernel matching method can be considered as a special case of our synchronous diffusion regularisation as shown in~\cref{sec:theoretical}.

\section{Background}
\label{sec:background}

\subsection{Diffusion Process}
\label{subsec:diffusion}
In the continuous case, the diffusion process of a scalar field $u$ is expressed by the heat equation, i.e.
\begin{equation}
\frac{\partial u}{\partial t} = \Delta u,
\end{equation}
where $\Delta$ is the Laplacian operator. The action of diffusion can be represented via the heat operator $h_t$, which is applied to some initial distribution $u_0$ and produces the diffused distribution $u_t$~at time $t$ \cite{sharp2020diffusionnet}. The solution of the diffusion equation can be expressed as
\begin{equation}
u_t=h_t\left(u_0\right)=\exp (t \Delta) u_0,
\end{equation}
where $\exp(\cdot)$ is the operator exponential. For discrete domains,  $\Delta$ is replaced by the Laplacian matrix $\mathbf{L} \in \mathbb{R}^{n \times n}$.
With $\mathbf{M} \in \mathbb{R}^{n \times n}$ being the mass matrix such that $\mathbf{M}^{-1} \mathbf{L} \approx-\Delta$, the diffusion process can be approximated (using the backward Euler method)~\cite{sharp2020diffusionnet} as
\begin{equation}
\label{eq:explicit_diffusion}
h_t(u):=(\mathbf{M}+t \mathbf{L})^{-1} \mathbf{M} u.
\end{equation}
However, computing the inverse of the (large) Laplacian matrix is both computationally expensive and numerically unstable~\cite{moler2003nineteen}. In practice,  spectral acceleration is often used~\cite{sharp2020diffusionnet,behmanesh2023tide}. Here, the diffusion process is approximated with the first $k$ eigenvectors $\mathbf{\Phi} \in \mathbb{R}^{n \times k}$ and the diagonal eigenvalue matrix $\mathbf{\Lambda} \in \mathbb{R}^{k \times k}$ of the Laplacian matrix, i.e.
\begin{equation}
h_t(u):=\mathbf{\Phi}\exp{\left(-t\mathbf{\Lambda}\right)}\mathbf{\Phi}^{\top} u, \quad\text{where}\quad\exp{\left(-t\mathbf{\Lambda}\right)} = \begin{pmatrix}
    e^{-\lambda_1 t} & & \\
    & \ddots & \\
    & & e^{-\lambda_k t}
  \end{pmatrix}.
  \end{equation}

\section{Synchronous Diffusion for Smooth Matching}
\label{sec:method}

\begin{figure}[bht!]
    \centering
    \includegraphics[width=\textwidth]{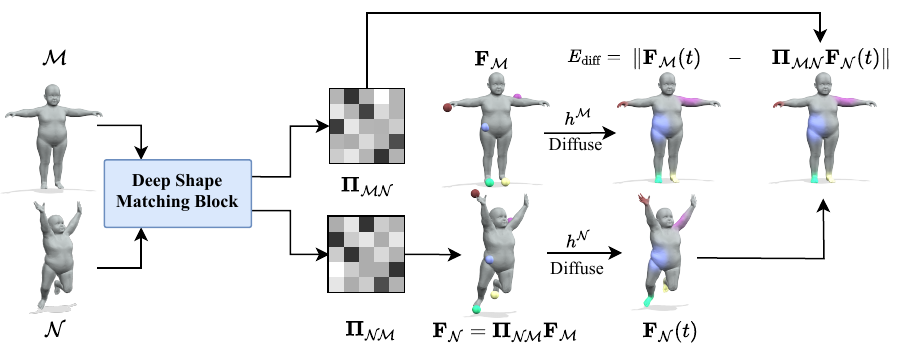}
    \caption{Illustration of our synchronous diffusion process for smooth matching. First, we transfer the function $\mathbf{F}_{\mathcal{M}}$ from shape $\mathcal{M}$ to shape $\mathcal{N}$ and perform synchronous diffusion on both shapes. Afterwards, the diffused function $\mathbf{F}_{\mathcal{N}}(t)$ from shape $\mathcal{N}$ is transferred back to shape $\mathcal{M}$. The difference between the two diffused functions is used to penalise spatially unsmooth pointwise correspondences.
    }
    \label{fig:sync_diffusion}
\end{figure}
{We propose a synchronous diffusion regulariser based on Definition~\ref{def:syncdiff} that can be applied in any  learning-based framework that outputs pointwise correspondences.} 
Typically, initial pointwise correspondences are obtained based on local feature matching~\cite{cao2023self,cao2023unsupervised,eisenberger2020deep} without considering \emph{neighbouring information} while using  regularisation in the \emph{low-frequency biased} spectral domain~\cite{cao2023self,sun2023spatially,attaiki2023understanding}. Therefore, the pointwise correspondences inevitably contain many local mismatches because this relies on high-frequency alignment. To address this limitation, we introduce a \emph{synchronous diffusion process} that encourages spatially smooth matchings. Specifically, consider an initial vector-valued function that is sampled from a random distribution $\mathbf{F}_{\mathcal{M}} \in \mathbb{R}^{n_\mathcal{M} \times h}$ defined on shape $\mathcal{M}$. We use the predicted soft pointwise correspondence $\mathbf{\Pi}_{\mathcal{NM}} \in \mathbb{R}^{n_\mathcal{N} \times n_\mathcal{M}}$ to transfer the function $\mathbf{F}_{\mathcal{M}}$ from shape $\mathcal{M}$ to shape $\mathcal{N}$, i.e.
\begin{equation}
    \mathbf{F}_{\mathcal{N}} := \mathbf{\Pi}_{\mathcal{NM}} \mathbf{F}_{\mathcal{M}}. 
\end{equation}
Afterwards, the diffusion process is applied synchronously to the initial function defined on both shapes, i.e.
\begin{equation}
    \mathbf{F}_{\mathcal{M}}(t) = h_t^{\mathcal{M}}\left(\mathbf{F}_{\mathcal{M}}\right), \quad \mathbf{F}_{\mathcal{N}}(t) = h_t^{\mathcal{N}}\left(\mathbf{F}_{\mathcal{N}}\right)=h_t^{\mathcal{N}}\left(\mathbf{\Pi}_{\mathcal{NM}} \mathbf{F}_{\mathcal{M}}\right).
\end{equation}
Next, we use the predicted pointwise map $\mathbf{\Pi}_{\mathcal{MN}} \in \mathbb{R}^{n_\mathcal{M} \times n_\mathcal{N}}$ to transfer the diffused function $\mathbf{F}_{\mathcal{N}}(t)$ from shape $\mathcal{N}$ back to shape $\mathcal{M}$, i.e.
\begin{equation}
    \mathbf{F}_{\mathcal{M}'}(t) := \mathbf{\Pi}_{\mathcal{MN}} \mathbf{F}_{\mathcal{N}}(t).
\end{equation}
Finally, the difference between the function $\mathbf{F}_{\mathcal{M}}(t)$ and the function $\mathbf{F}_{\mathcal{M}'}(t)$ is measured as a regularisation that penalises the non-smoothness of the matching, i.e.
\begin{equation}
    \label{eq:e_diff}
    E_{\mathrm{diff}} = \left\|\mathbf{F}_{\mathcal{M}}(t) - \mathbf{F}_{\mathcal{M}'}(t)\right\|^{2}_{F} = \left\|h_t^{\mathcal{M}}\left(\mathbf{F}_{\mathcal{M}}\right) - \mathbf{\Pi}_{\mathcal{MN}}h_t^{\mathcal{N}}\left(\mathbf{\Pi}_{\mathcal{NM}} \mathbf{F}_{\mathcal{M}}\right)\right\|^{2}_{F}.
\end{equation}
{Since diffusion exchanges information along neighbourhoods but without relying on exact discretisation of the mesh, the term penalises  correspondence that do not preserve local neighbourhoods while indicating at which areas respective violations occur.
In practice, the initial function $\mathbf{F}_{\mathcal{M}}$ is random sampled with unit norm along the feature dimension $h$, which ensures that $\mathbf{F}_{\mathcal{M}}$ is uniformly distributed on the entire shape $\mathcal{M}$ (i.e.\ with the same amount of heat in each vertex). Instead of manually fixing the diffusion time $t$~\cite{vestner2017efficient}, we randomly sample different diffusion times for different initial functions individually, i.e.
\begin{equation}
    \label{eq:t_i}
    t_i \sim \operatorname{Uniform}(0, T), \quad i \in \{1,\ldots h\},
\end{equation}
where $T$ is a pre-defined maximum diffusion time that controls the local/global information aggregation. To this end, our regularisation can be expressed as 
\begin{equation}
    \label{eq:l_diff}
    L_{\mathrm{diff}} = \sum_{i=1}^{h}\left\|h_{t_i}^{\mathcal{M}}\left(\mathbf{F}^i_{\mathcal{M}}\right) - \mathbf{\Pi}_{\mathcal{MN}}h_{t_i}^{\mathcal{N}}\left(\mathbf{\Pi}_{\mathcal{NM}} \mathbf{F}^i_{\mathcal{M}}\right)\right\|^{2}_{F},
\end{equation}
where $\mathbf{F}^i_{\mathcal{M}}$ denotes the $i$-th column of matrix $\mathbf{F}_{\mathcal{M}}$. As such, we have for each time $t_i$ a different scalar-valued initial function. In this case, our loss $L_{\mathrm{diff}}$ enables multiscale regularisation by using different diffusion times for each function.
The diffusion process on 3D shapes with a large diffusion time is analogous to performing several message passing operations in the graph and vice versa. %

\section{Theoretical Justification}
\label{sec:theoretical}

We will motivate the diffusion loss from two different perspectives: an approximation of a quadratic assignment problem and a continuous version of the Weisfeiler-Leman test. 

A common formulation for non-rigid shape correspondence is as a quadratic assignment problem of the form 
\begin{align}\label{eq:qap}
    \underset{\mathbf{\Pi}_\mathcal{NM} \in \mathcal{P}_n}{\arg \min} \Vert \mathbf{D_\M} - \mathbf{\Pi^\top_\mathcal{NM}} \mathbf{D_\N} \mathbf{\Pi}_\mathcal{NM} \Vert
\end{align}
where $\mathbf{D}_\bullet$ might be chosen as the heat kernel \cite{vestner2017efficient}. In that case, the formulation penalises the differences between heat diffusion based on \emph{initial Dirac functions} on all points. In \cite{talmon2019latentcommon} this is solved using a series of alternating diffusion steps  with the objective of  promoting neighbourhood preservation.  
The concept of synchronous diffusion is closely related: instead of using Dirac deltas, we sample from a random distribution in each iteration, which can be interpreted as a form of \emph{matrix sketching}~\cite{woodruff2014sketching}. In this case we decrease the dimension from $n$ (for Dirac functions in \cref{eq:qap}) to $h \ll n$.
Additionally, we show experimentally that using random functions instead of Dirac functions increases both the empirical performance and runtime, see Section~\ref{sec:ablation}.

Our loss can also be understood as a continuous version of the Weisfeiler-Leman test.
In the continuous domain, manifolds do not have the same discrete notion of neighbourhood (the meshing can only provide a noisy replacement) and the notion of discrete labels in each node is less meaningful. 
However, conceptually, heat diffusion provides both a weighted notion of neighbourhood as well as a message passing mechanism. 
In \cite{fey2019deep} the graph matching problem is solved through synchronous message passing aggregation of permuted random features. Our approach can be seen as replacing the neighbourhood message passing with heat diffusion, i.e.
\begin{align}\label{eq:passing}
    \mathbf{F}_\M^{(t)}(i) = \text{aggregate}( \{\!\{  \mathbf{F}_\M^{(t-1)}(j) \vert\ j \in \mathcal{N}(i) \}\!\}) \stackrel{\text{becomes}}{\rightsquigarrow} \mathbf{F}_\M^{(t)}(i) = h_{t}(\mathbf{F}^{(0)}_\M)(i).
\end{align}
Instead of a discrete number of message-passing steps, the influence of distant points is determined by the diffusion time parameter $t$, as was also proposed in \cite{behmanesh2023tide}; the update step becomes a simple integration.
Additionally, the Weisfeiler-Leman distance has been shown to be a lower bound for the Gromov-Wasserstein distance \cite{chen2022gromov,chen2023weisfeiler} which in turn is related to solving \cref{eq:qap}.

\section{Experimental Results}
\label{sec:experiment}
In this section we compare our method to previous methods on various shape matching datasets with different settings (including near-isometric, topologically noisy, non-isometric, and partial shape matching). To this end, we integrate our regularisation into the state-of-the-art unsupervised functional map framework of~\cite{cao2023unsupervised}. The details of~\cite{cao2023unsupervised} can be found in our supplementary material. 

\subsection{Near-isometric shape matching}
\noindent \textbf{Datasets.} Following prior works~\cite{sharma2020weakly,donati2020deep}, we evaluate our method on three standard benchmark datasets, namely the FAUST~\cite{bogo2014faust}, SCAPE~\cite{anguelov2005scape} and SHREC'19~\cite{melzi2019shrec} datasets. To evaluate the robustness against different discretisation, we choose the more challenging remeshed versions from~\cite{ren2018continuous}. The FAUST dataset contains 100 shapes, where the train/test split is 80/20. The SCAPE dataset consists of 71 shapes, where the last 20 shapes are used for evaluation. The SHREC'19 dataset is a more challenging dataset with significant variance in the mesh connectivity and shape geometry~\cite{melzi2019shrec} and used only for evaluation not training.

\noindent \textbf{Results.} We compare our method to state-of-the-art axiomatic, supervised and unsupervised methods. We use the mean geodesic error~\cite{kim2011blended} as our evaluation metric. The quantitative and qualitative results are summarised in~\cref{tab:near-iso} and~\cref{fig:near-iso}. Our
method outperforms the previous state of the art, even in comparison to supervised methods. Meanwhile, our method
achieves substantially better cross-dataset generalisation  compared to existing learning-based methods (see last column in~\cref{tab:near-iso} left).

\begin{figure}[!hbt]
\begin{tabular}{cc}
    \resizebox{0.55\linewidth}{!}{
\setlength{\tabcolsep}{2.5pt}
\small
\begin{tabular}{@{}lccc@{}}
\toprule
\multicolumn{1}{l}{Train}  & \multicolumn{1}{c}{\textbf{FAUST}}   & \multicolumn{1}{c}{\textbf{SCAPE}}  & \multicolumn{1}{c}{\textbf{FAUST + SCAPE}} \\ \cmidrule(lr){2-2} \cmidrule(lr){3-3} \cmidrule(lr){4-4}
\multicolumn{1}{l}{Test} & \multicolumn{1}{c}{\textbf{FAUST}} &  \multicolumn{1}{c}{\textbf{SCAPE}} &  \multicolumn{1}{c}{\textbf{SHREC'19}}
\\ \midrule
\multicolumn{4}{c}{Axiomatic Methods} \\
\multicolumn{1}{l}{BCICP~\cite{ren2018continuous}} & \multicolumn{1}{c}{6.1}  & \multicolumn{1}{c}{11.0} & \multicolumn{1}{c}{-}\\
\multicolumn{1}{l}{ZoomOut~\cite{melzi2019zoomout}} & \multicolumn{1}{c}{6.1} & \multicolumn{1}{c}{7.5} &  \multicolumn{1}{c}{-}\\
\multicolumn{1}{l}{Smooth Shells~\cite{eisenberger2020smooth}} & \multicolumn{1}{c}{2.5}  & \multicolumn{1}{c}{4.7} & \multicolumn{1}{c}{-}\\ 
\multicolumn{1}{l}{DiscreteOp~\cite{ren2021discrete}} & \multicolumn{1}{c}{5.6}  & \multicolumn{1}{c}{13.1} & \multicolumn{1}{c}{-}\\ 
\midrule
\multicolumn{4}{c}{Supervised Methods} \\ 
\multicolumn{1}{l}{FMNet~\cite{litany2017deep}} & \multicolumn{1}{c}{11.0} & \multicolumn{1}{c}{17.0} & \multicolumn{1}{c}{-} \\

\multicolumn{1}{l}{3D-CODED~\cite{groueix20183d}} & \multicolumn{1}{c}{2.5}  & \multicolumn{1}{c}{31.0} & \multicolumn{1}{c}{-} \\
\multicolumn{1}{l}{GeomFMaps~\cite{donati2020deep}}& \multicolumn{1}{c}{2.6} & \multicolumn{1}{c}{3.0} & \multicolumn{1}{c}{7.9}\\
\multicolumn{1}{l}{TransMatch~\cite{trappolini2021shape}}& \multicolumn{1}{c}{1.7}  & \multicolumn{1}{c}{12.0} & \multicolumn{1}{c}{10.9}\\
\midrule
\multicolumn{4}{c}{Unsupervised Methods} \\ 

\multicolumn{1}{l}{Deep Shells~\cite{eisenberger2020deep}} & \multicolumn{1}{c}{1.7}  & \multicolumn{1}{c}{2.5} &  \multicolumn{1}{c}{21.1} \\

\multicolumn{1}{l}{DUO-FMNet~\cite{donati2022deep}}  & \multicolumn{1}{c}{2.5}  & \multicolumn{1}{c}{2.6} & \multicolumn{1}{c}{6.4} \\
\multicolumn{1}{l}{AttnFMaps~\cite{li2022learning}}  & \multicolumn{1}{c}{1.9}  & \multicolumn{1}{c}{2.2} & \multicolumn{1}{c}{5.8}\\
\multicolumn{1}{l}{AttnFMaps-Fast~\cite{li2022learning}}  & \multicolumn{1}{c}{1.9}  & \multicolumn{1}{c}{2.1} &  \multicolumn{1}{c}{6.3}\\
\multicolumn{1}{l}{SSCDFM~\cite{sun2023spatially}}  & \multicolumn{1}{c}{1.7}  & \multicolumn{1}{c}{2.6} &  \multicolumn{1}{c}{3.8}\\
\multicolumn{1}{l}{URSSM~\cite{cao2023unsupervised}}  & \multicolumn{1}{c}{1.6}  & \multicolumn{1}{c}{1.9} &  \multicolumn{1}{c}{4.6}\\
\multicolumn{1}{l}{Ours}  & \multicolumn{1}{c}{\textbf{1.5}}  & \multicolumn{1}{c}{\textbf{1.8}} & \multicolumn{1}{c}{\textbf{3.4}} \\\hline
\end{tabular}
} & 
\hspace{-0.5cm}
\resizebox{0.45\linewidth}{!}{
\def\rowOnecolumnOne{11-5}
\def\rowOnecolumnTwo{11-5}
\def\rowTwocolumnOne{11-27}
\def\rowTwocolumnTwo{11-44}
\def\pathShrecNT{figs/ours/shrec19/}
\def\hspaceCols{-0.15cm}
\def\wspaceRows{0cm}
\def\height{3.8cm}
\def\width{3.4cm}
\def\heightT{\height}
\def\widthT{\width}
\begin{tabular}{cc}%
    \setlength{\tabcolsep}{0pt} 
    {\scriptsize Source} & \\
    \vspace{\wspaceRows}
    \hspace{\hspaceCols}
    \includegraphics[height=\heightT, width=\widthT]{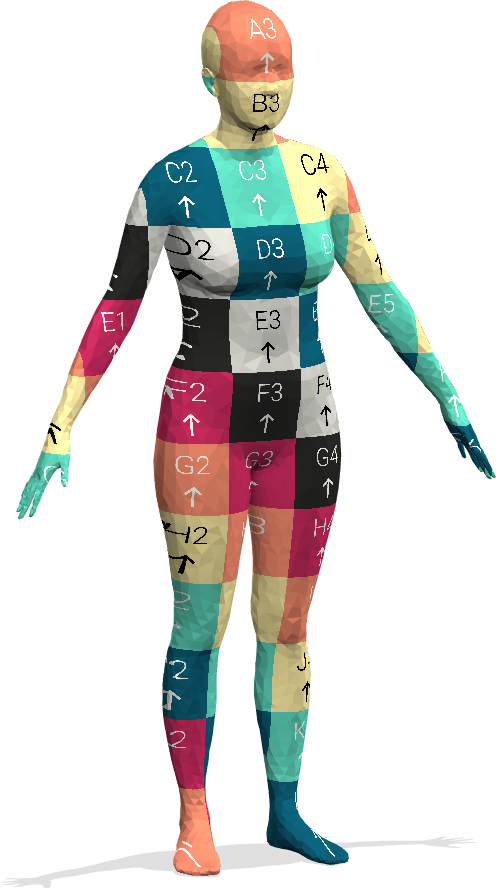}&
    \hspace{\hspaceCols}
    \includegraphics[height=\heightT, width=\widthT]{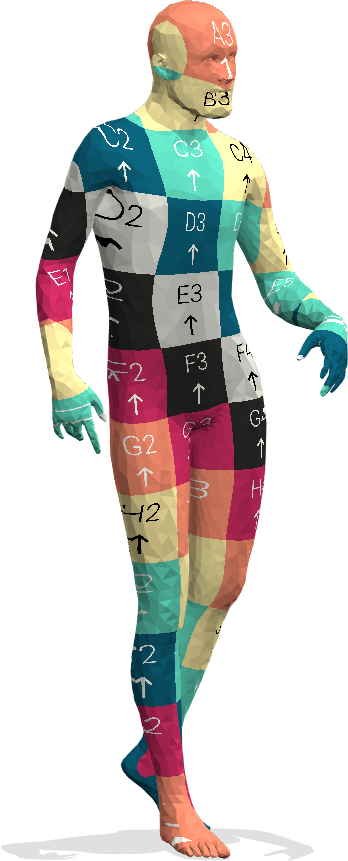}
    \\
    \hspace{\hspaceCols}
    \includegraphics[height=\heightT, width=\widthT]{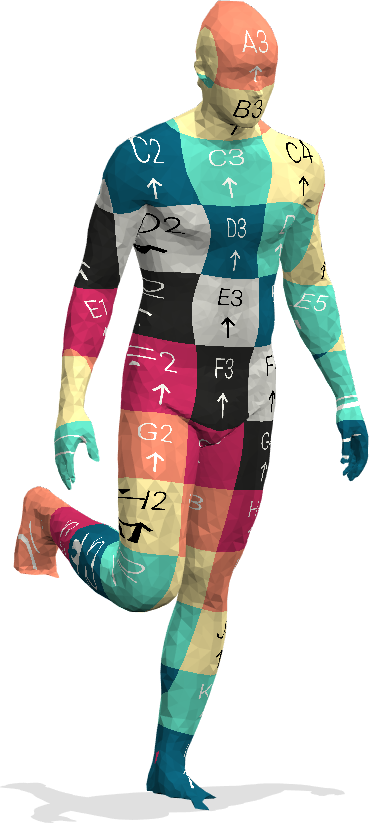}&
    \hspace{\hspaceCols}
    \includegraphics[height=\heightT, width=\widthT]{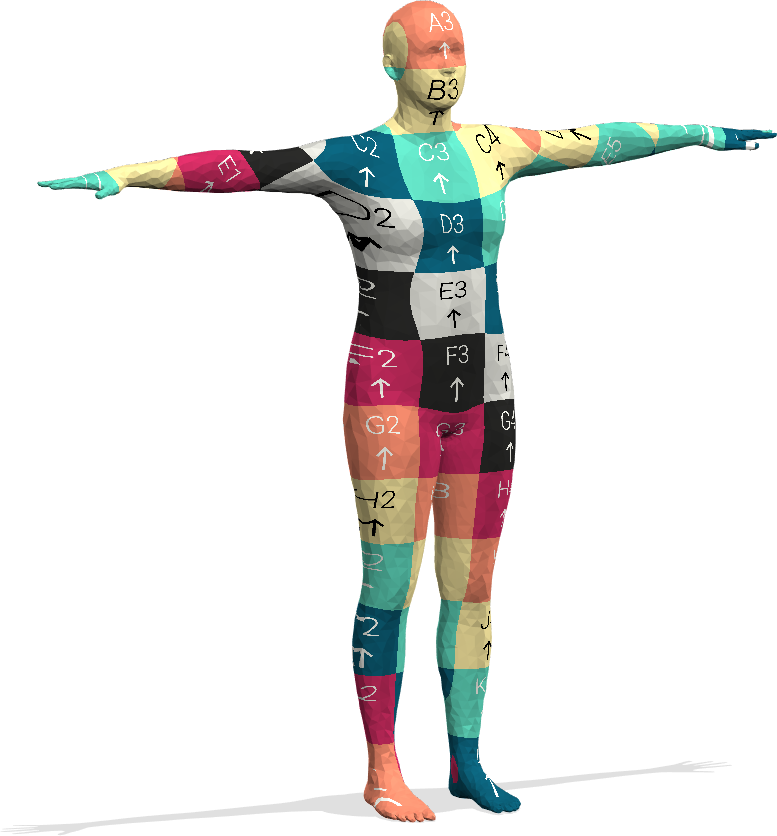}
\end{tabular}} \\
\end{tabular}
\caption{\textbf{Left: }Near-isometric shape matching and cross-dataset generalisation on FAUST, SCAPE and SHREC'19 datasets. The mean geodesic error~\cite{kim2011blended} is used as quantitative evaluation metric. The \textbf{best} results in each column are highlighted. \textbf{Right: } Qualitative results of our method on the challenging SHREC'19 dataset trained on FAUST and SCAPE datasets. Our method outperforms existing state-of-the-art methods and demonstrates superior cross-dataset generalisation ability.}
\label{tab:near-iso}
\end{figure}

\begin{figure}[!bht]
    \centering
    \begin{tabular}{ccc}
     \hspace{-1cm}
     \newcommand{\pckLineWidth}{1pt}
\newcommand{\plotWidth}{0.6\columnwidth}
\newcommand{\plotHeight}{0.4\columnwidth}
\newcommand{\pckTitle}{\textbf{FAUST}}
\definecolor{cPLOT0}{RGB}{28,213,227}
\definecolor{cPLOT1}{RGB}{80,150,80}
\definecolor{cPLOT2}{RGB}{90,130,213}
\definecolor{cPLOT3}{RGB}{247,179,43}
\definecolor{cPLOT4}{RGB}{124,42,43}
\definecolor{cPLOT5}{RGB}{242,64,0}

\pgfplotsset{%
    label style = {font=\normalsize},
    tick label style = {font=\normalsize},
    title style =  {font=\normalsize},
    legend style={  fill= gray!10,
                    fill opacity=0.6, 
                    font=\normalsize,
                    draw=gray!20, %
                    text opacity=1}
}
\begin{tikzpicture}[scale=0.55, transform shape]
	\begin{axis}[
		width=\plotWidth,
		height=\plotHeight,
		grid=major,
		title=\pckTitle,
		legend style={
			at={(0.97,0.03)},
			anchor=south east,
			legend columns=1},
		legend cell align={left},
        xlabel={\large Mean Geodesic Error},
		xmin=0,
        xmax=0.1,
        ylabel near ticks,
        xtick={0, 0.025, 0.05, 0.075, 0.1},
	ymin=0,
        ymax=1,
        ytick={0, 0.20, 0.40, 0.60, 0.80, 1.0}
	]

\addplot [color=cPLOT5, smooth, line width=\pckLineWidth]
table[row sep=crcr]{%
0.0 0.25808947368421054 \\
0.005263157894736842 0.26970736842105264 \\
0.010526315789473684 0.34548736842105265 \\
0.015789473684210527 0.4920557894736842 \\
0.021052631578947368 0.6559094736842105 \\
0.02631578947368421 0.7761673684210526 \\
0.031578947368421054 0.8456231578947369 \\
0.03684210526315789 0.8862315789473684 \\
0.042105263157894736 0.9158031578947369 \\
0.04736842105263158 0.9377410526315789 \\
0.05263157894736842 0.9530284210526315 \\
0.05789473684210526 0.9637242105263157 \\
0.06315789473684211 0.9721652631578948 \\
0.06842105263157895 0.9783347368421053 \\
0.07368421052631578 0.98296 \\
0.07894736842105263 0.9864778947368421 \\
0.08421052631578947 0.9891210526315789 \\
0.08947368421052632 0.9914873684210527 \\
0.09473684210526316 0.9931968421052632 \\
0.1 0.9947884210526315 \\
    };
\addlegendentry{\textcolor{black}{AttnFMaps: {0.82}}}    

\addplot [color=cPLOT3, smooth, line width=\pckLineWidth]
table[row sep=crcr]{%
0.0 0.312412 \\
0.005263157894736842 0.3258105 \\
0.010526315789473684 0.4023205 \\
0.015789473684210527 0.5468495 \\
0.021052631578947368 0.705658 \\
0.02631578947368421 0.8208325 \\
0.031578947368421054 0.8832595 \\
0.03684210526315789 0.921061 \\
0.042105263157894736 0.9465335 \\
0.04736842105263158 0.9657115 \\
0.05263157894736842 0.978877 \\
0.05789473684210526 0.9863015 \\
0.06315789473684211 0.9904765 \\
0.06842105263157895 0.993006 \\
0.07368421052631578 0.994677 \\
0.07894736842105263 0.9958235 \\
0.08421052631578947 0.9966505 \\
0.08947368421052632 0.997221 \\
0.09473684210526316 0.997665 \\
0.1 0.9979685 \\
    };
\addlegendentry{\textcolor{black}{URSSM: {0.85}}}  

\addplot [color=cPLOT1, smooth, line width=\pckLineWidth]
table[row sep=crcr]{%
0.0 0.326904 \\
0.005263157894736842 0.340823 \\
0.010526315789473684 0.4186815 \\
0.015789473684210527 0.564155 \\
0.021052631578947368 0.723681 \\
0.02631578947368421 0.837487 \\
0.031578947368421054 0.8979065 \\
0.03684210526315789 0.9334475 \\
0.042105263157894736 0.956158 \\
0.04736842105263158 0.972439 \\
0.05263157894736842 0.9830935 \\
0.05789473684210526 0.989045 \\
0.06315789473684211 0.992461 \\
0.06842105263157895 0.994471 \\
0.07368421052631578 0.9957675 \\
0.07894736842105263 0.99664 \\
0.08421052631578947 0.997234 \\
0.08947368421052632 0.9976495 \\
0.09473684210526316 0.997993 \\
0.1 0.9982205 \\
    };
\addlegendentry{\textcolor{black}{Ours: \textbf{0.86}}}  

\end{axis}
\end{tikzpicture}&
     \hspace{-1cm}
     \newcommand{\pckLineWidth}{1pt}
\newcommand{\plotWidth}{0.6\columnwidth}
\newcommand{\plotHeight}{0.4\columnwidth}
\newcommand{\pckTitle}{\textbf{SCAPE}}
\definecolor{cPLOT0}{RGB}{28,213,227}
\definecolor{cPLOT1}{RGB}{80,150,80}
\definecolor{cPLOT2}{RGB}{90,130,213}
\definecolor{cPLOT3}{RGB}{247,179,43}
\definecolor{cPLOT4}{RGB}{124,42,43}
\definecolor{cPLOT5}{RGB}{242,64,0}

\pgfplotsset{%
    label style = {font=\normalsize},
    tick label style = {font=\normalsize},
    title style =  {font=\normalsize},
    legend style={  fill= gray!10,
                    fill opacity=0.6, 
                    font=\normalsize,
                    draw=gray!20, %
                    text opacity=1}
}
\begin{tikzpicture}[scale=0.55, transform shape]
	\begin{axis}[
		width=\plotWidth,
		height=\plotHeight,
		grid=major,
		title=\pckTitle,
		legend style={
			at={(0.97,0.03)},
			anchor=south east,
			legend columns=1},
		legend cell align={left},
        xlabel={\large Mean Geodesic Error},
		xmin=0,
        xmax=0.1,
        ylabel near ticks,
        xtick={0, 0.025, 0.05, 0.075, 0.1},
	ymin=0,
        ymax=1,
        ytick={0, 0.20, 0.40, 0.60, 0.80, 1.0}
	]

\addplot [color=cPLOT5, smooth, line width=\pckLineWidth]
table[row sep=crcr]{%
0.0 0.18994526315789473 \\
0.005263157894736842 0.19249263157894736 \\
0.010526315789473684 0.22945157894736842 \\
0.015789473684210527 0.34391894736842105 \\
0.021052631578947368 0.5259463157894737 \\
0.02631578947368421 0.6822547368421052 \\
0.031578947368421054 0.77422 \\
0.03684210526315789 0.8329821052631579 \\
0.042105263157894736 0.8792747368421052 \\
0.04736842105263158 0.9162526315789473 \\
0.05263157894736842 0.9429273684210526 \\
0.05789473684210526 0.9598747368421052 \\
0.06315789473684211 0.9704663157894737 \\
0.06842105263157895 0.9777010526315789 \\
0.07368421052631578 0.98298 \\
0.07894736842105263 0.9869557894736842 \\
0.08421052631578947 0.9898021052631579 \\
0.08947368421052632 0.9920073684210526 \\
0.09473684210526316 0.9938705263157894 \\
0.1 0.995238947368421 \\
    };
\addlegendentry{\textcolor{black}{AttnFMaps: {0.78}}}

\addplot [color=cPLOT3, smooth, line width=\pckLineWidth]
table[row sep=crcr]{%
0.0 0.2509255 \\
0.005263157894736842 0.2536675 \\
0.010526315789473684 0.290562 \\
0.015789473684210527 0.404467 \\
0.021052631578947368 0.588076 \\
0.02631578947368421 0.738729 \\
0.031578947368421054 0.8219865 \\
0.03684210526315789 0.87358 \\
0.042105263157894736 0.911547 \\
0.04736842105263158 0.942448 \\
0.05263157894736842 0.964943 \\
0.05789473684210526 0.978441 \\
0.06315789473684211 0.985795 \\
0.06842105263157895 0.9904385 \\
0.07368421052631578 0.993671 \\
0.07894736842105263 0.995846 \\
0.08421052631578947 0.997343 \\
0.08947368421052632 0.9983675 \\
0.09473684210526316 0.999008 \\
0.1 0.9994695 \\
    };
\addlegendentry{\textcolor{black}{URSSM: {0.81}}}  

\addplot [color=cPLOT1, smooth, line width=\pckLineWidth]
table[row sep=crcr]{%
0.0 0.2625435 \\
0.005263157894736842 0.265402 \\
0.010526315789473684 0.3032245 \\
0.015789473684210527 0.41885 \\
0.021052631578947368 0.6047195 \\
0.02631578947368421 0.757951 \\
0.031578947368421054 0.8409745 \\
0.03684210526315789 0.8907065 \\
0.042105263157894736 0.926048 \\
0.04736842105263158 0.953287 \\
0.05263157894736842 0.9724355 \\
0.05789473684210526 0.9834915 \\
0.06315789473684211 0.9892725 \\
0.06842105263157895 0.9929285 \\
0.07368421052631578 0.995286 \\
0.07894736842105263 0.996897 \\
0.08421052631578947 0.9979965 \\
0.08947368421052632 0.9987425 \\
0.09473684210526316 0.9992315 \\
0.1 0.9995965 \\
    };
\addlegendentry{\textcolor{black}{Ours: \textbf{0.82}}} 

\end{axis}
\end{tikzpicture}&
     \hspace{-1cm}
     \newcommand{\pckLineWidth}{1pt}
\newcommand{\plotWidth}{0.6\columnwidth}
\newcommand{\plotHeight}{0.4\columnwidth}
\newcommand{\pckTitle}{\textbf{SHREC'19}}
\definecolor{cPLOT0}{RGB}{28,213,227}
\definecolor{cPLOT1}{RGB}{80,150,80}
\definecolor{cPLOT2}{RGB}{90,130,213}
\definecolor{cPLOT3}{RGB}{247,179,43}
\definecolor{cPLOT4}{RGB}{124,42,43}
\definecolor{cPLOT5}{RGB}{242,64,0}

\pgfplotsset{%
    label style = {font=\normalsize},
    tick label style = {font=\normalsize},
    title style =  {font=\normalsize},
    legend style={  fill= gray!10,
                    fill opacity=0.6, 
                    font=\normalsize,
                    draw=gray!20, %
                    text opacity=1}
}
\begin{tikzpicture}[scale=0.55, transform shape]
	\begin{axis}[
		width=\plotWidth,
		height=\plotHeight,
		grid=major,
		title=\pckTitle,
		legend style={
			at={(0.97,0.03)},
			anchor=south east,
			legend columns=1},
		legend cell align={left},
        xlabel={\large Mean Geodesic Error},
		xmin=0,
        xmax=0.1,
        ylabel near ticks,
        xtick={0, 0.025, 0.05, 0.075, 0.1},
	ymin=0,
        ymax=1,
        ytick={0, 0.20, 0.40, 0.60, 0.80, 1.0}
	]

\addplot [color=cPLOT5, smooth, line width=\pckLineWidth]
table[row sep=crcr]{%
0.0 0.08273666372582543 \\
0.005263157894736842 0.0867563906621049 \\
0.010526315789473684 0.12378170771564648 \\
0.015789473684210527 0.21680501844795372 \\
0.021052631578947368 0.3444028927298248 \\
0.02631578947368421 0.4566912324517737 \\
0.031578947368421054 0.545454416646043 \\
0.03684210526315789 0.6172611637040998 \\
0.042105263157894736 0.676746132352969 \\
0.04736842105263158 0.7270293221392448 \\
0.05263157894736842 0.7686236485197241 \\
0.05789473684210526 0.8019310369582507 \\
0.06315789473684211 0.8293578449616211 \\
0.06842105263157895 0.8517268153712166 \\
0.07368421052631578 0.8700538608459422 \\
0.07894736842105263 0.884449026783066 \\
0.08421052631578947 0.8961166726805925 \\
0.08947368421052632 0.9051814851689032 \\
0.09473684210526316 0.9122352534444682 \\
0.1 0.9177559429237504 \\
    };
\addlegendentry{\textcolor{black}{AttnFMaps: {0.63}}} 

\addplot [color=cPLOT3, smooth, line width=\pckLineWidth]
table[row sep=crcr]{%
0.0  0.08590200386528554 \\
0.002564102564102564  0.0861934116340183 \\
0.005128205128205128  0.09229833354429304 \\
0.007692307692307693  0.11138294475815516 \\
0.010256410256410256  0.1440328946001243 \\
0.01282051282051282  0.1936572288963155 \\
0.015384615384615385  0.2588485000765123 \\
0.017948717948717947  0.33116013352804596 \\
0.020512820512820513  0.4011296419793436 \\
0.023076923076923078  0.46355655388823924 \\
0.02564102564102564  0.5173862281727553 \\
0.028205128205128206  0.565500626266939 \\
0.03076923076923077  0.6088330087072831 \\
0.03333333333333333  0.6476521129660872 \\
0.035897435897435895  0.6827443149508716 \\
0.038461538461538464  0.7140850551738339 \\
0.041025641025641026  0.7424829075410851 \\
0.04358974358974359  0.7681164006370353 \\
0.046153846153846156  0.7912415198922406 \\
0.04871794871794872  0.8119886195111906 \\
0.05128205128205128  0.8302264006937111 \\
0.05384615384615385  0.846344036861902 \\
0.05641025641025641  0.8603835625237329 \\
0.05897435897435897  0.8729655770440579 \\
0.06153846153846154  0.8839724782601304 \\
0.0641025641025641  0.8936678083679842 \\
0.06666666666666667  0.9025597598459931 \\
0.06923076923076923  0.9104287141974621 \\
0.07179487179487179  0.9173521282685372 \\
0.07435897435897436  0.9234967231975697 \\
0.07692307692307693  0.9289172855342728 \\
0.07948717948717948  0.9336752139981523 \\
0.08205128205128205  0.9379164486224961 \\
0.08461538461538462  0.9415625879654731 \\
0.08717948717948718  0.9449702168980612 \\
0.08974358974358974  0.9478809885004922 \\
0.09230769230769231  0.9504110880419854 \\
0.09487179487179487  0.9526833129615531 \\
0.09743589743589744  0.954680188239028 \\
0.1  0.9563913233218785 \\
    };
\addlegendentry{\textcolor{black}{URSSM: {0.69}}}       

\addplot [color=cPLOT1, smooth, line width=\pckLineWidth]
table[row sep=crcr]{%
0.0 0.08756827065311235 \\
0.005263157894736842 0.09550004061761444 \\
0.010526315789473684 0.15369374696548635 \\
0.015789473684210527 0.2789291874399001 \\
0.021052631578947368 0.4282664591075082 \\
0.02631578947368421 0.5474716007307392 \\
0.031578947368421054 0.6412321683949619 \\
0.03684210526315789 0.7166038195671107 \\
0.042105263157894736 0.7773625754968101 \\
0.04736842105263158 0.8258307246749174 \\
0.05263157894736842 0.863695314994323 \\
0.05789473684210526 0.8924530583174484 \\
0.06315789473684211 0.9145414838086854 \\
0.06842105263157895 0.9318785930058358 \\
0.07368421052631578 0.9454902168224935 \\
0.07894736842105263 0.9558982443744604 \\
0.08421052631578947 0.9642073803150037 \\
0.08947368421052632 0.9707128107955951 \\
0.09473684210526316 0.9756757165230566 \\
0.1 0.9794885770043849 \\
    };
\addlegendentry{\textcolor{black}{Ours: \textbf{0.71}}}  
\end{axis}
\end{tikzpicture}
    \end{tabular}
    \caption{Near-isometric shape matching and cross-dataset generalisation on FAUST, SCAPE and SHREC'19 datasets where we evaluate the performance on the SHREC'19 dataset trained on FAUST and SCAPE datasets. Proportion of correct keypoints (PCK) curves and corresponding area under curve (scores in the legend) of our method in comparison to the existing unsupervised state-of-the-art methods.}
    \label{fig:near-iso}
\end{figure}
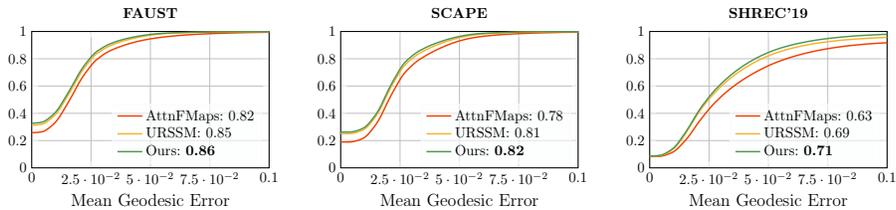

\subsection{Matching with topological noise}
\textbf{Datasets.} The topology of meshes can undergo significant inconsistencies, e.g.~due to self-intersections of separate parts of real-world scanned objects. Such topological noise presents a great challenge, both for matching methods based on functional maps as it distorts the intrinsic shape geometry~\cite{lahner2016shrec}, as well as methods solely based on spatial registrations~\cite{eisenberger2023g}. To evaluate our method for matching with topologically noisy shapes, we use the TOPKIDS dataset~\cite{lahner2016shrec}.

\begin{figure}[!bht]
    \begin{tabular}{cc}
    \resizebox{0.60\linewidth}{!}{
    \setlength{\tabcolsep}{4.5pt}
    \small
    \begin{tabular}{@{}lcc@{}}
    \toprule
    \multicolumn{1}{c}{\textbf{Geo. error ($\times$100)}}      & \multicolumn{1}{c}{\textbf{TOPKIDS}} & \multicolumn{1}{c}{\textbf{Fully intrinsic}}
    \\ \midrule
    \multicolumn{3}{c}{Axiomatic Methods} \\
    \multicolumn{1}{l}{ZoomOut~\cite{melzi2019zoomout}}  & \multicolumn{1}{c}{33.7} & \multicolumn{1}{c}{\cmark}  \\ 
    \multicolumn{1}{l}{Smooth Shells~\cite{eisenberger2020smooth}}  & \multicolumn{1}{c}{11.8} & \multicolumn{1}{c}{\xmark}\\
    \multicolumn{1}{l}{DiscreteOp~\cite{ren2021discrete}}  & \multicolumn{1}{c}{35.5} & \multicolumn{1}{c}{\cmark}
    \\\midrule
    \multicolumn{3}{c}{Unsupervised Methods} \\
    \multicolumn{1}{l}{UnsupFMNet~\cite{halimi2019unsupervised}}  & \multicolumn{1}{c}{38.5} & \multicolumn{1}{c}{\cmark}\\
    \multicolumn{1}{l}{SURFMNet~\cite{roufosse2019unsupervised}}  & \multicolumn{1}{c}{48.6} & \multicolumn{1}{c}{\cmark}\\
    \multicolumn{1}{l}{WSupFMNet~\cite{sharma2020weakly}}  & \multicolumn{1}{c}{47.9} & \multicolumn{1}{c}{\cmark}\\
    \multicolumn{1}{l}{Deep Shells~\cite{eisenberger2020deep}}  & \multicolumn{1}{c}{13.7} & \multicolumn{1}{c}{\xmark} \\
    \multicolumn{1}{l}{NeuroMorph~\cite{eisenberger2021neuromorph}}  & \multicolumn{1}{c}{13.8} & \multicolumn{1}{c}{\xmark} \\
    \multicolumn{1}{l}{AttnFMaps~\cite{li2022learning}}  & \multicolumn{1}{c}{23.4} & \multicolumn{1}{c}{\cmark} \\
    \multicolumn{1}{l}{AttnFMaps-Fast~\cite{li2022learning}}  & \multicolumn{1}{c}{28.5} & \multicolumn{1}{c}{\cmark}\\
    \multicolumn{1}{l}{URSSM~\cite{cao2023unsupervised}}  & \multicolumn{1}{c}{{9.2}} & \multicolumn{1}{c}{\cmark} \\
    \multicolumn{1}{l}{Ours}  & \multicolumn{1}{c}{\textbf{5.4}} & \multicolumn{1}{c}{\cmark} \\
    \hline
    \end{tabular}
    }
         &
    \hspace{-1cm}
    \resizebox{0.40\linewidth}{!}{
\def\filename{kid00-kid17}
\def\pathOurs{figs/ours/topkids/}
\def\pathURSSM{figs/urssm/topkids/}
\def\pathAttn{figs/attnfmaps/topkids/}
\def\hspaceCols{-0.2cm}
\def\wspaceRows{0cm}
\def\height{3.8cm}
\def\width{3.4cm}
\def\heightT{\height}
\def\widthT{\width}
\begin{tabular}{cc}%
    \setlength{\tabcolsep}{0pt} 
    {\scriptsize Source} & {\scriptsize AttnFMaps} \\
    \vspace{\wspaceRows}
    \hspace{\hspaceCols}
    \includegraphics[height=\heightT, width=\widthT]{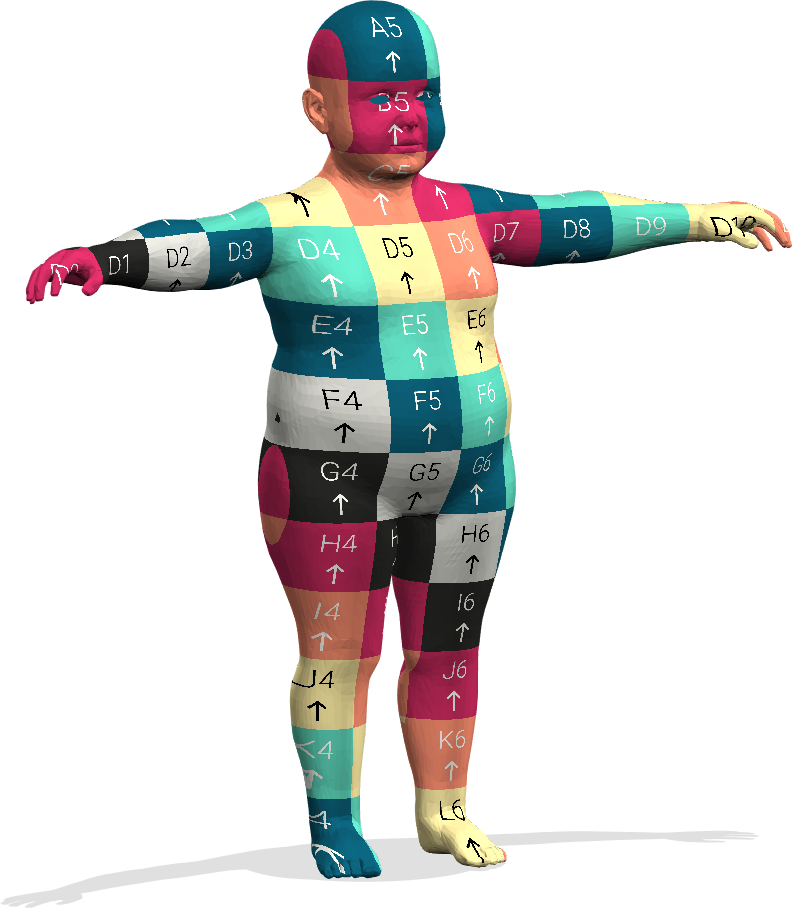}&
    \hspace{\hspaceCols}
    \includegraphics[height=\heightT, width=\widthT]{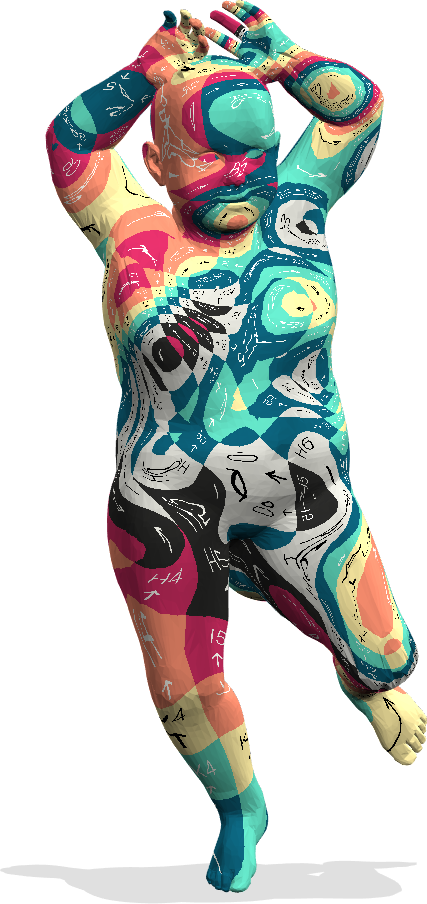}
    \\
    {\scriptsize URSSM} & {\scriptsize Ours} \\
    \vspace{\wspaceRows}
    \hspace{\hspaceCols}
    \includegraphics[height=\heightT, width=\widthT]{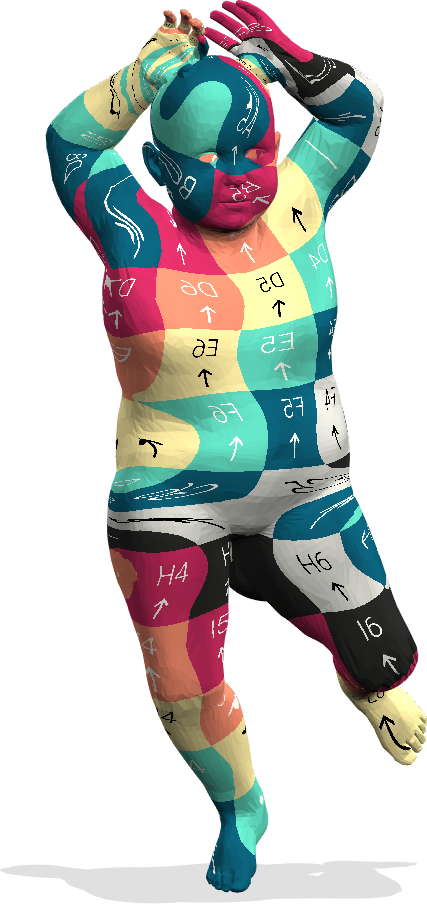}&
    \hspace{\hspaceCols}
    \includegraphics[height=\heightT, width=\widthT]{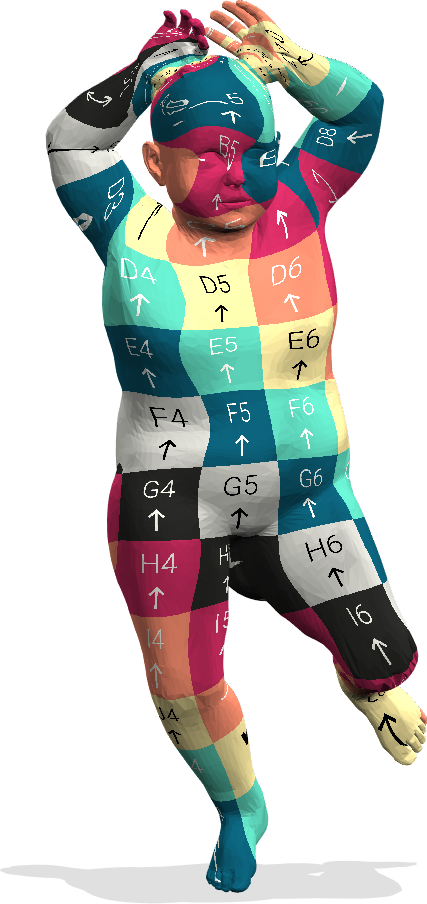}
\end{tabular}} \\
    \end{tabular}
    \caption{{Results on the TOPKIDS dataset.} We distinguish fully intrinsic methods (i.e.\ methods solely based on functional maps) from the methods that rely on additional extrinsic information (e.g.\ rigid alignment).  Our method substantially outperforms all existing methods, even compared to methods relying on extrinsic information.}
    \label{fig:topkids}
\end{figure}

\noindent \textbf{Results.} We compare our method with prominent axiomatic methods and unsupervised methods. The quantitative results are summarised in~\cref{fig:topkids}. Our method substantially outperforms existing methods, even in comparison to methods relying on additional extrinsic information.~\cref{fig:noniso-pck} left shows PCK curves of our method in comparison to existing state-of-the-art unsupervised methods and demonstrates the superior performance of our method even under large topological noise. The main reason is the relative robustness of the diffusion process w.r.t. small topological noise~\cite{vestner2017efficient} when choosing small diffusion times $t$ and the multiscale regularisation based on randomly sampled diffusion time (see comparison to fixed diffusion time in~\cref{sec:ablation}). By choosing a small maximum diffusion time $T$ in~\cref{eq:t_i}, we ensure our regularisation to focus more on the local neighbourhood and thus to be less effected by the global distortion caused from topological noisy parts. More details about the choice of diffusion time can be found in the supplementary materials.   

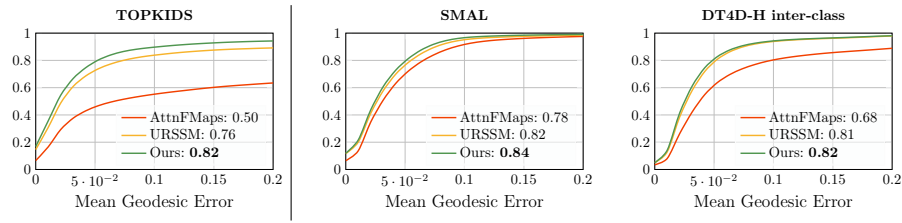
\begin{figure}[!bht]
    \centering
    \begin{tabular}{c|cc}
     \hspace{-1cm}
     \newcommand{\pckLineWidth}{1pt}
\newcommand{\plotWidth}{0.6\columnwidth}
\newcommand{\plotHeight}{0.4\columnwidth}
\newcommand{\pckTitle}{\textbf{TOPKIDS}}
\definecolor{cPLOT0}{RGB}{28,213,227}
\definecolor{cPLOT1}{RGB}{80,150,80}
\definecolor{cPLOT2}{RGB}{90,130,213}
\definecolor{cPLOT3}{RGB}{247,179,43}
\definecolor{cPLOT4}{RGB}{124,42,43}
\definecolor{cPLOT5}{RGB}{242,64,0}

\pgfplotsset{%
    label style = {font=\normalsize},
    tick label style = {font=\normalsize},
    title style =  {font=\normalsize},
    legend style={  fill= gray!10,
                    fill opacity=0.6, 
                    font=\normalsize,
                    draw=gray!20, %
                    text opacity=1}
}
\begin{tikzpicture}[scale=0.55, transform shape]
	\begin{axis}[
		width=\plotWidth,
		height=\plotHeight,
		grid=major,
		title=\pckTitle,
		legend style={
			at={(0.97,0.03)},
			anchor=south east,
			legend columns=1},
		legend cell align={left},
        xlabel={\large Mean Geodesic Error},
		xmin=0,
        xmax=0.2,
        ylabel near ticks,
        xtick={0, 0.05, 0.1, 0.15, 0.2},
	ymin=0,
        ymax=1,
        ytick={0, 0.20, 0.40, 0.60, 0.80, 1.0}
	]

\addplot [color=cPLOT5, smooth, line width=\pckLineWidth]
table[row sep=crcr]{%
0.0 0.06531700633935275 \\
0.010526315789473684 0.16553915459815935 \\
0.021052631578947368 0.2928293328585914 \\
0.031578947368421054 0.37656451974952204 \\
0.042105263157894736 0.4298994527567283 \\
0.05263157894736842 0.46736665299203517 \\
0.06315789473684211 0.4941521599467463 \\
0.07368421052631578 0.5145363912905498 \\
0.08421052631578947 0.5312052510584939 \\
0.09473684210526316 0.5456061860936738 \\
0.10526315789473684 0.5583274635622673 \\
0.11578947368421053 0.5697406206218604 \\
0.12631578947368421 0.5801591417491659 \\
0.1368421052631579 0.5905737926977468 \\
0.14736842105263157 0.5999860673565905 \\
0.15789473684210525 0.6084191867980463 \\
0.16842105263157894 0.6156177192262738 \\
0.17894736842105263 0.6220847878755041 \\
0.18947368421052632 0.6285015442013112 \\
0.2 0.6345622440844318 \\
    };
\addlegendentry{\textcolor{black}{AttnFMaps: {0.50}}}    

\addplot [color=cPLOT3, smooth, line width=\pckLineWidth]
table[row sep=crcr]{%
0.0 0.1457277097056342 \\
0.010526315789473684 0.30846098472827477 \\
0.021052631578947368 0.48728259271013136 \\
0.031578947368421054 0.609193222543017 \\
0.042105263157894736 0.6856911752184716 \\
0.05263157894736842 0.737915366931645 \\
0.06315789473684211 0.7737454815663387 \\
0.07368421052631578 0.7998188756356769 \\
0.08421052631578947 0.8187479197789354 \\
0.09473684210526316 0.832680563188408 \\
0.10526315789473684 0.843435789864776 \\
0.11578947368421053 0.8532583034684542 \\
0.12631578947368421 0.8613972893268211 \\
0.1368421052631579 0.8683016881719597 \\
0.14736842105263157 0.8742308019784354 \\
0.15789473684210525 0.8792426834271206 \\
0.16842105263157894 0.8829967567902286 \\
0.17894736842105263 0.8861354717360848 \\
0.18947368421052632 0.8891309900691214 \\
0.2 0.8914879289125572 \\
    };
\addlegendentry{\textcolor{black}{URSSM: {0.76}}}  

\addplot [color=cPLOT1, smooth, line width=\pckLineWidth]
table[row sep=crcr]{%
0.0 0.16844178864179948 \\
0.010526315789473684 0.35656343609947905 \\
0.021052631578947368 0.5445341465868894 \\
0.031578947368421054 0.6672691244881689 \\
0.042105263157894736 0.746340746015651 \\
0.05263157894736842 0.8023809339515299 \\
0.06315789473684211 0.8398016920421385 \\
0.07368421052631578 0.8649075414302633 \\
0.08421052631578947 0.8810229656405533 \\
0.09473684210526316 0.8923316278745752 \\
0.10526315789473684 0.9015039514524781 \\
0.11578947368421053 0.9096855092768183 \\
0.12631578947368421 0.9165434659772588 \\
0.1368421052631579 0.9224067867454119 \\
0.14736842105263157 0.9272445101514788 \\
0.15789473684210525 0.9316913455063355 \\
0.16842105263157894 0.9351164536778308 \\
0.17894736842105263 0.9382125966577136 \\
0.18947368421052632 0.9406043671096731 \\
0.2 0.942640081118946 \\
    };
\addlegendentry{\textcolor{black}{Ours: \textbf{0.82}}}  

\end{axis}
\end{tikzpicture}&
     \hspace{-1cm}
     \newcommand{\pckLineWidth}{1pt}
\newcommand{\plotWidth}{0.6\columnwidth}
\newcommand{\plotHeight}{0.4\columnwidth}
\newcommand{\pckTitle}{\textbf{SMAL}}
\definecolor{cPLOT0}{RGB}{28,213,227}
\definecolor{cPLOT1}{RGB}{80,150,80}
\definecolor{cPLOT2}{RGB}{90,130,213}
\definecolor{cPLOT3}{RGB}{247,179,43}
\definecolor{cPLOT4}{RGB}{124,42,43}
\definecolor{cPLOT5}{RGB}{242,64,0}

\pgfplotsset{%
    label style = {font=\normalsize},
    tick label style = {font=\normalsize},
    title style =  {font=\normalsize},
    legend style={  fill= gray!10,
                    fill opacity=0.6, 
                    font=\normalsize,
                    draw=gray!20, %
                    text opacity=1}
}
\begin{tikzpicture}[scale=0.55, transform shape]
	\begin{axis}[
		width=\plotWidth,
		height=\plotHeight,
		grid=major,
		title=\pckTitle,
		legend style={
			at={(0.97,0.03)},
			anchor=south east,
			legend columns=1},
		legend cell align={left},
        xlabel={\large Mean Geodesic Error},
		xmin=0,
        xmax=0.2,
        ylabel near ticks,
        xtick={0, 0.05, 0.1, 0.15, 0.2},
	ymin=0,
        ymax=1,
        ytick={0, 0.20, 0.40, 0.60, 0.80, 1.0}
	]

\addplot [color=cPLOT5, smooth, line width=\pckLineWidth]
table[row sep=crcr]{%
0.0 0.06488340934619913 \\
0.010526315789473684 0.13704104694753083 \\
0.021052631578947368 0.3482007280994979 \\
0.031578947368421054 0.5091134238269884 \\
0.042105263157894736 0.6313163984788405 \\
0.05263157894736842 0.7207291821737424 \\
0.06315789473684211 0.788068912316791 \\
0.07368421052631578 0.8386001001475146 \\
0.08421052631578947 0.877629210593983 \\
0.09473684210526316 0.9058721630509805 \\
0.10526315789473684 0.9262048152007687 \\
0.11578947368421053 0.9395975152589625 \\
0.12631578947368421 0.9487271792234507 \\
0.1368421052631579 0.9551447402254672 \\
0.14736842105263157 0.9603266974327049 \\
0.15789473684210525 0.9645085328389114 \\
0.16842105263157894 0.9678350543367934 \\
0.17894736842105263 0.9708245929815539 \\
0.18947368421052632 0.9734906822211095 \\
0.2 0.9757561814023359 \\
    };
\addlegendentry{\textcolor{black}{AttnFMaps: {0.78}}}    

\addplot [color=cPLOT3, smooth, line width=\pckLineWidth]
table[row sep=crcr]{%
0.0 0.11681666238107483 \\
0.010526315789473684 0.2009649010028285 \\
0.021052631578947368 0.4125514271020828 \\
0.031578947368421054 0.5778947030084854 \\
0.042105263157894736 0.6992298791463101 \\
0.05263157894736842 0.7843635896117254 \\
0.06315789473684211 0.8472923630753407 \\
0.07368421052631578 0.893088840318848 \\
0.08421052631578947 0.9245872975057855 \\
0.09473684210526316 0.944853432759064 \\
0.10526315789473684 0.9571213679609154 \\
0.11578947368421053 0.9645358704037027 \\
0.12631578947368421 0.9692600925687838 \\
0.1368421052631579 0.9726851375674981 \\
0.14736842105263157 0.9756203394188737 \\
0.15789473684210525 0.9781679094883003 \\
0.16842105263157894 0.9803272049370018 \\
0.17894736842105263 0.9822512213936745 \\
0.18947368421052632 0.9838518899460016 \\
0.2 0.98525777834919 \\
    };
\addlegendentry{\textcolor{black}{URSSM: {0.82}}}  

\addplot [color=cPLOT1, smooth, line width=\pckLineWidth]
table[row sep=crcr]{%
0.0 0.11935523270763693 \\
0.010526315789473684 0.21620982257649782 \\
0.021052631578947368 0.4373193623039342 \\
0.031578947368421054 0.60819876574955 \\
0.042105263157894736 0.7283980457701209 \\
0.05263157894736842 0.8127314219593725 \\
0.06315789473684211 0.8745860118282335 \\
0.07368421052631578 0.9171933659038313 \\
0.08421052631578947 0.9447441501671381 \\
0.09473684210526316 0.9612406788377474 \\
0.10526315789473684 0.9706049112882489 \\
0.11578947368421053 0.9759533299048598 \\
0.12631578947368421 0.9795204422730779 \\
0.1368421052631579 0.9821586526099254 \\
0.14736842105263157 0.9843481614811005 \\
0.15789473684210525 0.9861339676009256 \\
0.16842105263157894 0.9876214965286706 \\
0.17894736842105263 0.9888255335561841 \\
0.18947368421052632 0.9898360761121111 \\
0.2 0.9906916945230136 \\
    };
\addlegendentry{\textcolor{black}{Ours: \textbf{0.84}}}  

\end{axis}
\end{tikzpicture}&
     \hspace{-1cm}
     \newcommand{\pckLineWidth}{1pt}
\newcommand{\plotWidth}{0.6\columnwidth}
\newcommand{\plotHeight}{0.4\columnwidth}
\newcommand{\pckTitle}{\textbf{DT4D-H inter-class}}
\definecolor{cPLOT0}{RGB}{28,213,227}
\definecolor{cPLOT1}{RGB}{80,150,80}
\definecolor{cPLOT2}{RGB}{90,130,213}
\definecolor{cPLOT3}{RGB}{247,179,43}
\definecolor{cPLOT4}{RGB}{124,42,43}
\definecolor{cPLOT5}{RGB}{242,64,0}

\pgfplotsset{%
    label style = {font=\normalsize},
    tick label style = {font=\normalsize},
    title style =  {font=\normalsize},
    legend style={  fill= gray!10,
                    fill opacity=0.6, 
                    font=\normalsize,
                    draw=gray!20, %
                    text opacity=1}
}
\begin{tikzpicture}[scale=0.55, transform shape]
	\begin{axis}[
		width=\plotWidth,
		height=\plotHeight,
		grid=major,
		title=\pckTitle,
		legend style={
			at={(0.97,0.03)},
			anchor=south east,
			legend columns=1},
		legend cell align={left},
        xlabel={\large Mean Geodesic Error},
		xmin=0,
        xmax=0.2,
        ylabel near ticks,
        xtick={0, 0.05, 0.1, 0.15, 0.2},
	ymin=0,
        ymax=1,
        ytick={0, 0.20, 0.40, 0.60, 0.80, 1.0}
	]

\addplot [color=cPLOT5, smooth, line width=\pckLineWidth]
table[row sep=crcr]{%
0.0 0.03456135231465077 \\
0.010526315789473684 0.08081747504012339 \\
0.021052631578947368 0.26366498530545884 \\
0.031578947368421054 0.4222855743377035 \\
0.042105263157894736 0.5487909831794402 \\
0.05263157894736842 0.6365265648123059 \\
0.06315789473684211 0.6959819288409029 \\
0.07368421052631578 0.7381741251016112 \\
0.08421052631578947 0.7694500489818038 \\
0.09473684210526316 0.7939294870458762 \\
0.10526315789473684 0.8120624465889905 \\
0.11578947368421053 0.8260296600454384 \\
0.12631578947368421 0.8372674406486441 \\
0.1368421052631579 0.8469622944327491 \\
0.14736842105263157 0.8553124413781603 \\
0.15789473684210525 0.8624852533505638 \\
0.16842105263157894 0.8689187527356859 \\
0.17894736842105263 0.8752861787940055 \\
0.18947368421052632 0.8819866811180357 \\
0.2 0.8887820205515142 \\
    };
\addlegendentry{\textcolor{black}{AttnFMaps: {0.68}}}    

\addplot [color=cPLOT3, smooth, line width=\pckLineWidth]
table[row sep=crcr]{%
0.0 0.050371427975905124 \\
0.010526315789473684 0.12944817308293557 \\
0.021052631578947368 0.3864891510515455 \\
0.031578947368421054 0.5803202576234445 \\
0.042105263157894736 0.7183634866706964 \\
0.05263157894736842 0.8076292806970007 \\
0.06315789473684211 0.8629943514600746 \\
0.07368421052631578 0.897993205077433 \\
0.08421052631578947 0.9198287721199742 \\
0.09473684210526316 0.9337159472247119 \\
0.10526315789473684 0.942982262334035 \\
0.11578947368421053 0.9494709965191654 \\
0.12631578947368421 0.9543339516851825 \\
0.1368421052631579 0.95824395022615 \\
0.14736842105263157 0.9615212497655127 \\
0.15789473684210525 0.9646511870271172 \\
0.16842105263157894 0.9679835129332806 \\
0.17894736842105263 0.9716687162598745 \\
0.18947368421052632 0.9755318173291369 \\
0.2 0.9789688809221085 \\
    };
\addlegendentry{\textcolor{black}{URSSM: {0.81}}}  

\addplot [color=cPLOT1, smooth, line width=\pckLineWidth]
table[row sep=crcr]{%
0.0 0.0504584488400692 \\
0.010526315789473684 0.1416884131979907 \\
0.021052631578947368 0.41568605790274504 \\
0.031578947368421054 0.6125830085249182 \\
0.042105263157894736 0.7456655272317986 \\
0.05263157894736842 0.8269095608312317 \\
0.06315789473684211 0.8763701356900181 \\
0.07368421052631578 0.9067878358380057 \\
0.08421052631578947 0.9257863142755904 \\
0.09473684210526316 0.9382403026450175 \\
0.10526315789473684 0.9467947141338559 \\
0.11578947368421053 0.9528558475936386 \\
0.12631578947368421 0.9573865602267754 \\
0.1368421052631579 0.9611137628446964 \\
0.14736842105263157 0.9643077933176314 \\
0.15789473684210525 0.9673428934697876 \\
0.16842105263157894 0.970753902911812 \\
0.17894736842105263 0.974391062384059 \\
0.18947368421052632 0.9781210788502824 \\
0.2 0.9809179398461763 \\
    };
\addlegendentry{\textcolor{black}{Ours: \textbf{0.82}}}  

\end{axis}
\end{tikzpicture}
    \end{tabular}
    \vspace{-2mm}
    \caption{\textbf{Left}: Shape matching with topological noise on the TOPKIDS dataset. Our method significantly outperforms existing unsupervised methods. \textbf{Right}: Non-isometric shape matching on SMAL and DT4D-H datasets. Our method demonstrates  state-of-the-art performance in both challenging scenarios.}
    \vspace{-3mm}
    \label{fig:noniso-pck}
\end{figure}

\subsection{Non-isometric shape matching}
\noindent \textbf{Datasets.} In the context of non-isometric shape matching, we consider the SMAL~\cite{zuffi20173d} dataset and the DT4D-H~\cite{magnet2022smooth} dataset. The SMAL dataset contains 49 animal shapes of 8 species. Similarly to previous work~\cite{donati2022deep,cao2023unsupervised}, five species are used for training and three different species are used for testing (i.e.\ 29/20 shapes for the train/test split). The DT4D-H dataset based on DeformingThings4D~\cite{li20214dcomplete} is firstly introduced in~\cite{magnet2022smooth}. Following prior evaluations~\cite{li2022learning,cao2023unsupervised}, nine classes of shapes are used for evaluation, resulting in 198/95 shapes for train/test split. 

\begin{figure}[!bht]
    \begin{tabular}{cc}
    \resizebox{0.56\linewidth}{!}{
    \setlength{\tabcolsep}{3.5pt}
    \small
    \begin{tabular}{@{}lccc@{}}
    \toprule
    \multicolumn{1}{l}{\multirow{2}{*}{\textbf{Geo. error ($\times$100)}}}  & \multicolumn{1}{c}{\multirow{2}{*}{\textbf{SMAL}}}   & \multicolumn{2}{c}{\textbf{DT4D-H}}\\ \cmidrule(lr){3-4}
    &  & \multicolumn{1}{c}{\textbf{intra-class}} & \multicolumn{1}{c}{\textbf{inter-class}}
    \\ \midrule
    \multicolumn{4}{c}{Axiomatic Methods} \\
    \multicolumn{1}{l}{ZoomOut~\cite{melzi2019zoomout}}  & \multicolumn{1}{c}{38.4} & \multicolumn{1}{c}{4.0} & \multicolumn{1}{c}{29.0} \\
    \multicolumn{1}{l}{Smooth Shells~\cite{eisenberger2020smooth}}  & \multicolumn{1}{c}{36.1} & \multicolumn{1}{c}{1.1} & \multicolumn{1}{c}{6.3} \\
    \multicolumn{1}{l}{DiscreteOp~\cite{ren2021discrete}}  & \multicolumn{1}{c}{38.1} & \multicolumn{1}{c}{3.6} & \multicolumn{1}{c}{27.6} \\
    \midrule
    \multicolumn{4}{c}{Supervised Methods} \\ 
    \multicolumn{1}{l}{FMNet~\cite{litany2017deep}}  & \multicolumn{1}{c}{42.0} & \multicolumn{1}{c}{9.6} & \multicolumn{1}{c}{38.0} \\
    \multicolumn{1}{l}{GeomFMaps~\cite{donati2020deep}}  & \multicolumn{1}{c}{8.4} & \multicolumn{1}{c}{2.1} & \multicolumn{1}{c}{4.1} \\
    \midrule
    \multicolumn{4}{c}{Unsupervised Methods} \\
    \multicolumn{1}{l}{WSupFMNet~\cite{sharma2020weakly}}  & \multicolumn{1}{c}{7.6} & \multicolumn{1}{c}{3.3} & \multicolumn{1}{c}{22.6} \\
    \multicolumn{1}{l}{Deep Shells~\cite{eisenberger2020deep}}  & \multicolumn{1}{c}{29.3} & \multicolumn{1}{c}{3.4} & \multicolumn{1}{c}{31.1} \\
    \multicolumn{1}{l}{DUO-FMNet~\cite{donati2022deep}}  & \multicolumn{1}{c}{6.7} & \multicolumn{1}{c}{2.6} & \multicolumn{1}{c}{15.8} \\
    \multicolumn{1}{l}{AttnFMaps~\cite{li2022learning}}  & \multicolumn{1}{c}{5.4} & \multicolumn{1}{c}{1.7} & \multicolumn{1}{c}{11.6} \\
    \multicolumn{1}{l}{AttnFMaps-Fast~\cite{li2022learning}}  & \multicolumn{1}{c}{5.8} & \multicolumn{1}{c}{1.2} & \multicolumn{1}{c}{14.6} \\
    \multicolumn{1}{l}{URSSM~\cite{cao2023unsupervised}}  & \multicolumn{1}{c}{3.9} & \multicolumn{1}{c}{\textbf{0.9}} & \multicolumn{1}{c}{4.1} \\
    \multicolumn{1}{l}{Ours}  & \multicolumn{1}{c}{\textbf{3.6}} & \multicolumn{1}{c}{\textbf{0.9}} & \multicolumn{1}{c}{\textbf{3.9}} \\
    \hline
    \end{tabular}
    }
         &
    \hspace{-1cm}
    \resizebox{0.44\linewidth}{!}{
    \centering
\def\filename{Strafing024-Standing2HMagicAttack01043}
\def\pathOurs{figs/ours/dt4d/}
\def\pathURSSM{figs/urssm/dt4d/}
\def\pathAttn{figs/attnfmaps/dt4d/}
\def\hspaceCols{-0.2cm}
\def\wspaceRows{0cm}
\def\height{3.8cm}
\def\width{3.4cm}
\def\heightT{\height}
\def\widthT{\width}
\begin{tabular}{cc}%
    \setlength{\tabcolsep}{0pt} 
    {\scriptsize Source} & {\scriptsize AttnFMaps} \\
    \vspace{\wspaceRows}
    \hspace{\hspaceCols}
    \includegraphics[height=\heightT, width=\widthT]{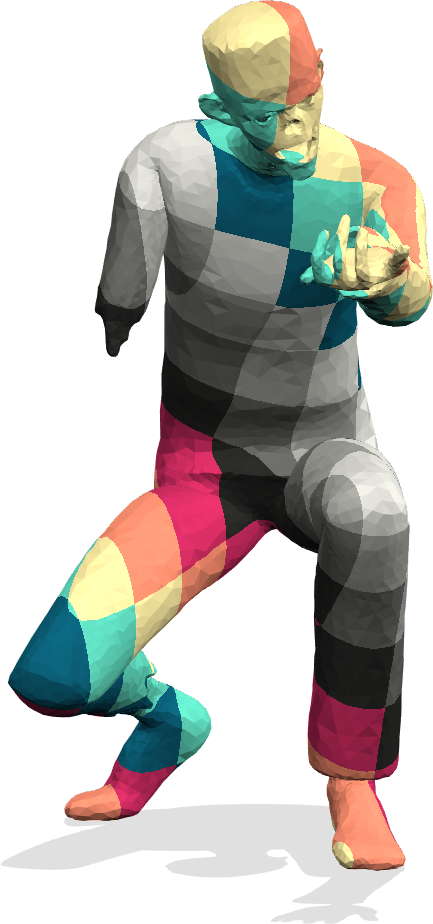}&
    \hspace{\hspaceCols}
    \begin{overpic}[height=\heightT, width=\widthT]{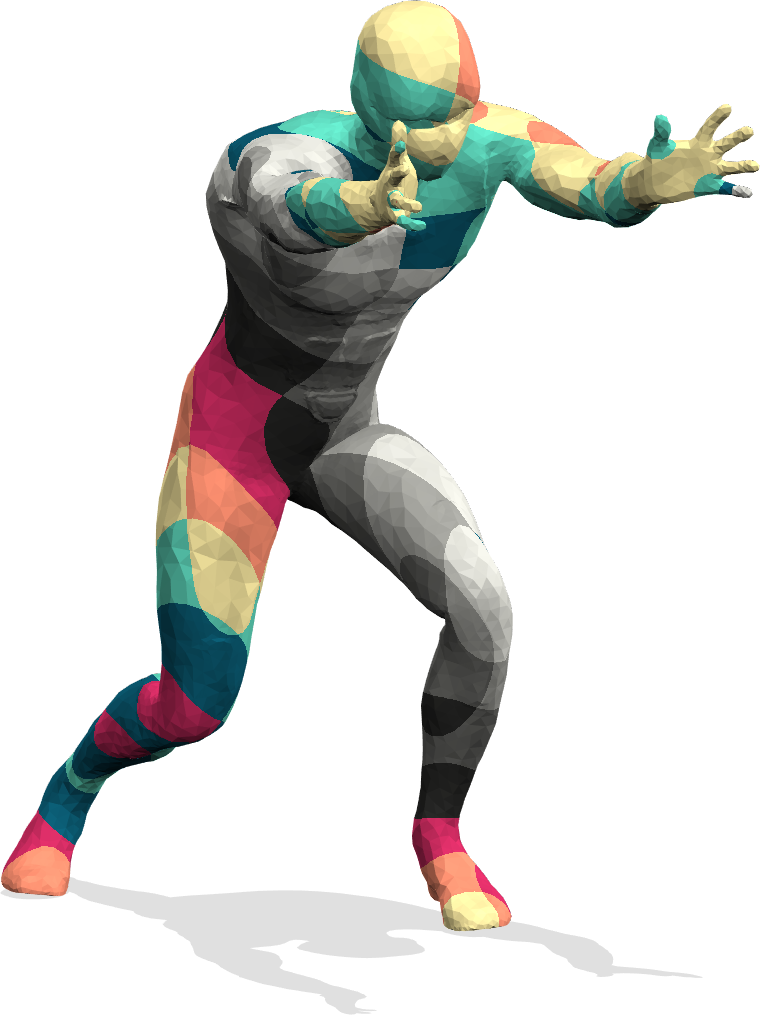}
       \put(30,72){\includegraphics[height=0.7cm]{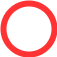}}
       \put(35,4){\includegraphics[height=0.8cm]{figs/red_circle.png}}
    \end{overpic}
    \\
    {\scriptsize URSSM} & {\scriptsize Ours} \\
    \vspace{\wspaceRows}
    \hspace{\hspaceCols}
    \begin{overpic}[height=\heightT, width=\widthT]{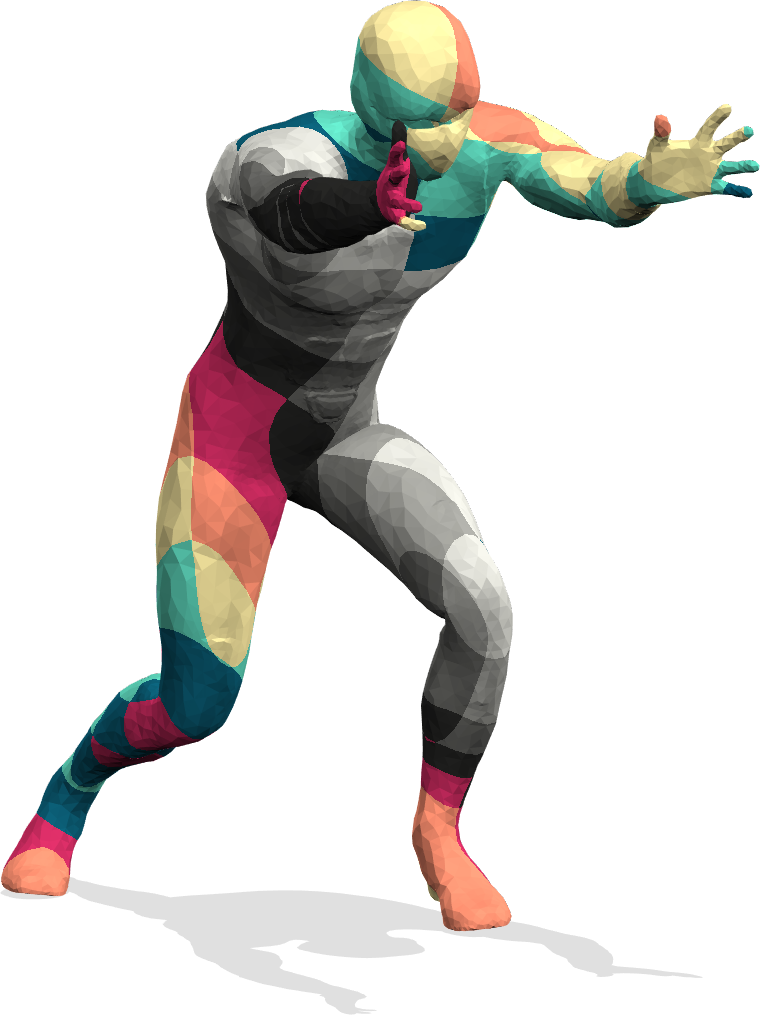}
       \put(30,72){\includegraphics[height=0.7cm]{figs/red_circle.png}}
    \end{overpic}&
    \hspace{\hspaceCols}
    \begin{overpic}[height=\heightT, width=\widthT]{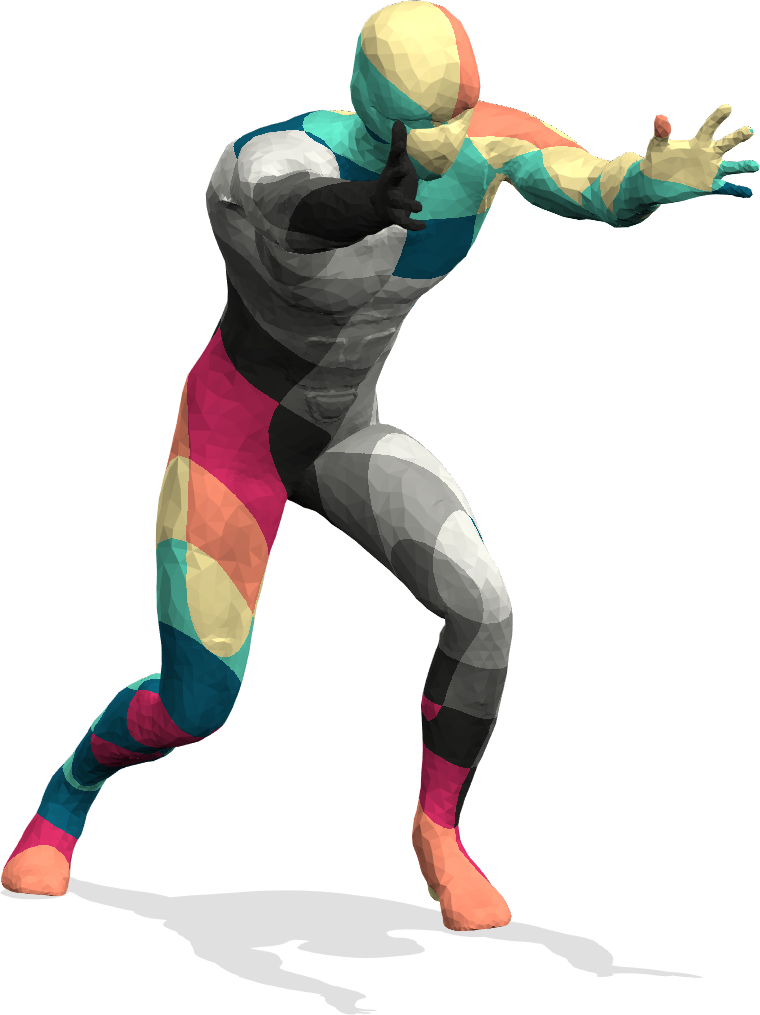}
       \put(30,72){\includegraphics[height=0.7cm]{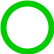}}
    \end{overpic}
\end{tabular}} \\
    \end{tabular}
    \vspace{-3mm}
    \caption{Results on SMAL and DT4D-H datasets. Our method outperforms existing methods for challenging non-isometric shape matching on both SMAL and DT4D-H inter-class datasets and shows comparable performance on near-isometric shape matching on the DT4D-H intra-class dataset.}
    \vspace{-5mm}
    \label{fig:noniso}
\end{figure}
\noindent \textbf{Results.}~\cref{fig:noniso} summarises the qualitative and quantitative matching results on the SMAL and DT4D-H datasets. In the context of non-isometric inter-class shape matching, our approach outperforms the existing state-of-the-art methods on both challenging datasets, even in comparison to supervised methods. Meanwhile, our method demonstrates comparable and near-perfect matching results for intra-class matching on the DT4D-H dataset.~\cref{fig:noniso-pck} right shows the PCK curves and the corresponding AUC of our method compared to existing state-of-the-art methods. Compared to the results on topological noisy dataset, the performance improvement is less significant -- one of the main reasons is that large non-isometry leads to inconsistent diffusion processes on two shapes.

\subsection{Partial shape matching}
\textbf{Datasets.} We evaluate our method on the SHREC’16 partial dataset~\cite{cosmo2016shrec}. The dataset contains 200 training shapes and 400 test shapes, with eight different classes (including humans and animals). Each class has a complete template shape to which the other partial shapes are matched. The dataset is divided into two subsets, namely CUTS (missing a large part) with a 120/200 train/test split and HOLES (missing many small parts) with a 80/200 train/test split.

\begin{figure}[!bht]
    \centering
\def\hspaceCols{0.4cm}
\def\wspaceRows{-0.0cm}
\def\height{2.6cm}
\def\width{3.6cm}
\def\heightT{\height}
\def\widthT{\width}
\def\filename{cat-cuts_cat_shape_16}
\def\pathOurs{figs/ours/shrec16/}
\def\pathURSSM{figs/urssm/shrec16/}
\def\pathDPFM{figs/dpfm/shrec16/}
\def\pathDPFMUnsup{figs/dpfm-unsup/shrec16/}
\begin{tabular}{ccccc}%
    \setlength{\tabcolsep}{0pt} 
    {\small Source}&
    \hspace{\hspaceCols}{\small DPFM}&
    \hspace{\hspaceCols}{\small DPFM-unsup}&
    \hspace{\hspaceCols}{\small URSSM}&
    \hspace{\hspaceCols}{\small Ours}
    \vspace{\wspaceRows}\\
    \includegraphics[height=\heightT, width=\widthT]{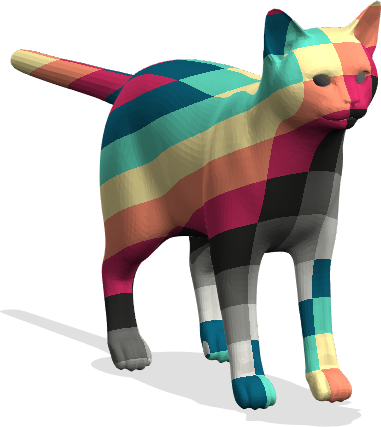}&
    \hspace{\hspaceCols}
    \includegraphics[height=\heightT, width=\widthT]{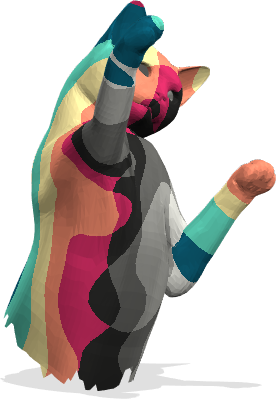}&
    \hspace{\hspaceCols}
    \includegraphics[height=\heightT, width=\widthT]{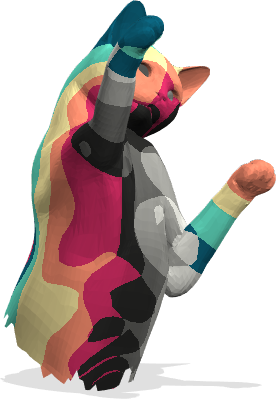}&
    \hspace{\hspaceCols}
    \includegraphics[height=\heightT, width=\widthT]{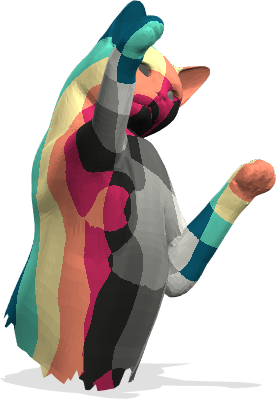}&
    \hspace{\hspaceCols}
    \includegraphics[height=\heightT, width=\widthT]{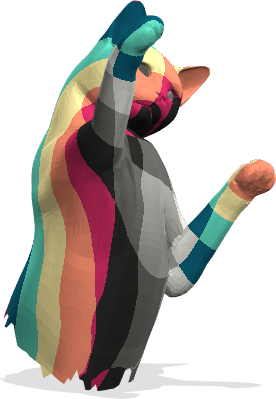}\\
\end{tabular}
    \vspace{-3mm}
    \caption{Qualitative comparison between our method and existing state-of-the-art methods on the SHREC'16 dataset. Our method leads to  spatially smoother matchings.}
    \vspace{-6mm}
    \label{fig:shrec16-plot}
\end{figure}

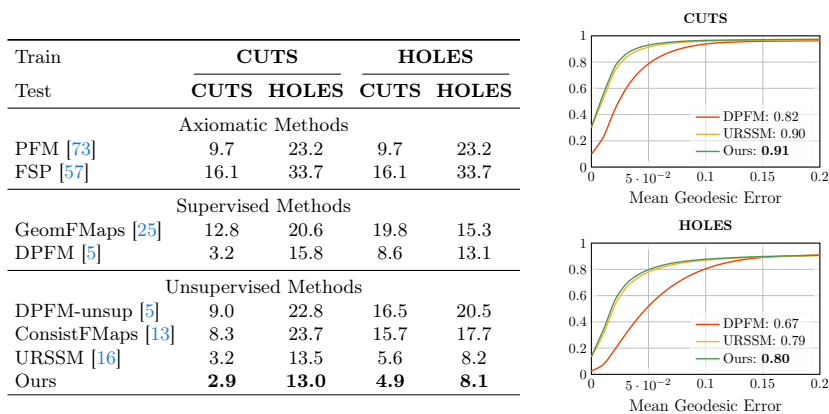
\begin{figure}[!bht]
    \begin{tabular}{cc}
    \resizebox{0.58\linewidth}{!}{
    \setlength{\tabcolsep}{3.5pt}
    \small
    \begin{tabular}{@{}lcccc@{}}
    \toprule
    \multicolumn{1}{l}{Train}  & \multicolumn{2}{c}{\textbf{CUTS}} & \multicolumn{2}{c}{\textbf{HOLES}}\\ \cmidrule(lr){2-3} \cmidrule(lr){4-5}
    \multicolumn{1}{l}{Test} & \multicolumn{1}{c}{\textbf{CUTS}} & \multicolumn{1}{c}{\textbf{HOLES}} & \multicolumn{1}{c}{\textbf{CUTS}} & \multicolumn{1}{c}{\textbf{HOLES}}
    \\ \midrule
    \multicolumn{5}{c}{Axiomatic Methods} \\
    \multicolumn{1}{l}{PFM~\cite{rodola2017partial}}  & \multicolumn{1}{c}{9.7} & \multicolumn{1}{c}{23.2} & \multicolumn{1}{c}{9.7} & \multicolumn{1}{c}{23.2} \\
    \multicolumn{1}{l}{FSP~\cite{litany2017fully}}  & \multicolumn{1}{c}{16.1} & \multicolumn{1}{c}{33.7}  & \multicolumn{1}{c}{16.1} & \multicolumn{1}{c}{33.7} \\
    \midrule
    \multicolumn{5}{c}{Supervised Methods} \\ 
    \multicolumn{1}{l}{GeomFMaps~\cite{donati2020deep}}  & \multicolumn{1}{c}{12.8} & \multicolumn{1}{c}{20.6} & \multicolumn{1}{c}{19.8} & \multicolumn{1}{c}{15.3}\\
    \multicolumn{1}{l}{DPFM~\cite{attaiki2021dpfm}}  & \multicolumn{1}{c}{3.2} & \multicolumn{1}{c}{15.8} & \multicolumn{1}{c}{8.6} & \multicolumn{1}{c}{13.1}\\
    \midrule
    \multicolumn{5}{c}{Unsupervised Methods} \\
    \multicolumn{1}{l}{DPFM-unsup~\cite{attaiki2021dpfm}}  & \multicolumn{1}{c}{9.0} & \multicolumn{1}{c}{22.8} & \multicolumn{1}{c}{16.5} & \multicolumn{1}{c}{20.5} \\
    \multicolumn{1}{l}{ConsistFMaps~\cite{cao2022unsupervised}}  & \multicolumn{1}{c}{8.3} & \multicolumn{1}{c}{23.7} & \multicolumn{1}{c}{15.7} & \multicolumn{1}{c}{17.7} \\
    \multicolumn{1}{l}{URSSM~\cite{cao2023unsupervised}}  & \multicolumn{1}{c}{3.2} & \multicolumn{1}{c}{{13.5}}  & \multicolumn{1}{c}{{5.6}}  & \multicolumn{1}{c}{{8.2}} \\
    \multicolumn{1}{l}{Ours}  & \multicolumn{1}{c}{\textbf{2.9}} & \multicolumn{1}{c}{\textbf{13.0}}  & \multicolumn{1}{c}{\textbf{4.9}}  & \multicolumn{1}{c}{\textbf{8.1}} \\
    \hline
    \end{tabular}
    }
         &
    \hspace{-1cm}
    \resizebox{0.42\linewidth}{!}{
    \centering
    \begin{tabular}{c}
       \newcommand{\pckLineWidth}{1pt}
\newcommand{\plotWidth}{0.6\columnwidth}
\newcommand{\plotHeight}{0.4\columnwidth}
\newcommand{\pckTitle}{\textbf{CUTS}}
\definecolor{cPLOT0}{RGB}{28,213,227}
\definecolor{cPLOT1}{RGB}{80,150,80}
\definecolor{cPLOT2}{RGB}{90,130,213}
\definecolor{cPLOT3}{RGB}{247,179,43}
\definecolor{cPLOT4}{RGB}{124,42,43}
\definecolor{cPLOT5}{RGB}{242,64,0}

\pgfplotsset{%
    label style = {font=\normalsize},
    tick label style = {font=\normalsize},
    title style =  {font=\normalsize},
    legend style={  fill= gray!10,
                    fill opacity=0.6, 
                    font=\normalsize,
                    draw=gray!20, %
                    text opacity=1}
}
\begin{tikzpicture}[scale=0.55, transform shape]
	\begin{axis}[
		width=\plotWidth,
		height=\plotHeight,
		grid=major,
		title=\pckTitle,
		legend style={
			at={(0.97,0.03)},
			anchor=south east,
			legend columns=1},
		legend cell align={left},
        xlabel={\large Mean Geodesic Error},
		xmin=0,
        xmax=0.2,
        ylabel near ticks,
        xtick={0, 0.05, 0.1, 0.15, 0.2},
	ymin=0,
        ymax=1,
        ytick={0, 0.20, 0.40, 0.60, 0.80, 1.0}
	]

\addplot [color=cPLOT5, smooth, line width=\pckLineWidth]
table[row sep=crcr]{%
0.0 0.09820953899901268 \\
0.010526315789473684 0.2295535239614187 \\
0.021052631578947368 0.4490357047922837 \\
0.031578947368421054 0.6150508847877268 \\
0.042105263157894736 0.7275171831092884 \\
0.05263157894736842 0.8033685064935064 \\
0.06315789473684211 0.8551836504139135 \\
0.07368421052631578 0.8906309333940913 \\
0.08421052631578947 0.9150563435102909 \\
0.09473684210526316 0.931423017771702 \\
0.10526315789473684 0.9422823156375788 \\
0.11578947368421053 0.949004851143009 \\
0.12631578947368421 0.9532175609478241 \\
0.1368421052631579 0.9562068048910154 \\
0.14736842105263157 0.9580342902711324 \\
0.15789473684210525 0.9593598105111263 \\
0.16842105263157894 0.9602723665223665 \\
0.17894736842105263 0.9610579479000532 \\
0.18947368421052632 0.9617877553732816 \\
0.2 0.9624594155844156 \\
    };
\addlegendentry{\textcolor{black}{DPFM: {0.82}}}    

\addplot [color=cPLOT3, smooth, line width=\pckLineWidth]
table[row sep=crcr]{%
0.0 0.30081240031897927 \\
0.010526315789473684 0.523318238778765 \\
0.021052631578947368 0.7392498765854029 \\
0.031578947368421054 0.8435755582137161 \\
0.042105263157894736 0.8925260594668489 \\
0.05263157894736842 0.919430441634389 \\
0.06315789473684211 0.9356464076858814 \\
0.07368421052631578 0.9457165641376167 \\
0.08421052631578947 0.9525720076706918 \\
0.09473684210526316 0.9573353364471785 \\
0.10526315789473684 0.9603566207184628 \\
0.11578947368421053 0.9624736557302347 \\
0.12631578947368421 0.9640732987772461 \\
0.1368421052631579 0.9652884578871421 \\
0.14736842105263157 0.9662745879851143 \\
0.15789473684210525 0.9669972753854332 \\
0.16842105263157894 0.9678516841345789 \\
0.17894736842105263 0.9686693058403585 \\
0.18947368421052632 0.9694786207944103 \\
0.2 0.9702392819169134 \\
    };
\addlegendentry{\textcolor{black}{URSSM: {0.90}}}  

\addplot [color=cPLOT1, smooth, line width=\pckLineWidth]
table[row sep=crcr]{%
0.0 0.3041101807549176 \\
0.010526315789473684 0.5575456159337738 \\
0.021052631578947368 0.7727023524720893 \\
0.031578947368421054 0.8687687495253285 \\
0.042105263157894736 0.9117099092428039 \\
0.05263157894736842 0.9341405122655123 \\
0.06315789473684211 0.9469127363864206 \\
0.07368421052631578 0.9549596054530265 \\
0.08421052631578947 0.9604610484544696 \\
0.09473684210526316 0.9639166571732362 \\
0.10526315789473684 0.9661915204678363 \\
0.11578947368421053 0.9677033492822966 \\
0.12631578947368421 0.9688461209842789 \\
0.1368421052631579 0.9698013974329763 \\
0.14736842105263157 0.9706534328244855 \\
0.15789473684210525 0.9714224006987164 \\
0.16842105263157894 0.972078634085213 \\
0.17894736842105263 0.9726161995898838 \\
0.18947368421052632 0.9731478317004633 \\
0.2 0.9737257442849548 \\
    };
\addlegendentry{\textcolor{black}{Ours: \textbf{0.91}}}  

\end{axis}
\end{tikzpicture}    \\
       \newcommand{\pckLineWidth}{1pt}
\newcommand{\plotWidth}{0.6\columnwidth}
\newcommand{\plotHeight}{0.4\columnwidth}
\newcommand{\pckTitle}{\textbf{HOLES}}
\definecolor{cPLOT0}{RGB}{28,213,227}
\definecolor{cPLOT1}{RGB}{80,150,80}
\definecolor{cPLOT2}{RGB}{90,130,213}
\definecolor{cPLOT3}{RGB}{247,179,43}
\definecolor{cPLOT4}{RGB}{124,42,43}
\definecolor{cPLOT5}{RGB}{242,64,0}

\pgfplotsset{%
    label style = {font=\normalsize},
    tick label style = {font=\normalsize},
    title style =  {font=\normalsize},
    legend style={  fill= gray!10,
                    fill opacity=0.6, 
                    font=\normalsize,
                    draw=gray!20, %
                    text opacity=1}
}
\begin{tikzpicture}[scale=0.55, transform shape]
	\begin{axis}[
		width=\plotWidth,
		height=\plotHeight,
		grid=major,
		title=\pckTitle,
		legend style={
			at={(0.97,0.03)},
			anchor=south east,
			legend columns=1},
		legend cell align={left},
        xlabel={\large Mean Geodesic Error},
		xmin=0,
        xmax=0.2,
        ylabel near ticks,
        xtick={0, 0.05, 0.1, 0.15, 0.2},
	ymin=0,
        ymax=1,
        ytick={0, 0.20, 0.40, 0.60, 0.80, 1.0}
	]

\addplot [color=cPLOT5, smooth, line width=\pckLineWidth]
table[row sep=crcr]{%
0.0 0.025942907493282673 \\
0.010526315789473684 0.07528458314250837 \\
0.021052631578947368 0.19621478735278905 \\
0.031578947368421054 0.32684963805656314 \\
0.042105263157894736 0.4429562515998823 \\
0.05263157894736842 0.5427140272454511 \\
0.06315789473684211 0.6240629260910979 \\
0.07368421052631578 0.6905988740311462 \\
0.08421052631578947 0.7444385044431017 \\
0.09473684210526316 0.7867607237823916 \\
0.10526315789473684 0.8196455966886137 \\
0.11578947368421053 0.845340359174661 \\
0.12631578947368421 0.8644849641539301 \\
0.1368421052631579 0.879182984740388 \\
0.14736842105263157 0.8888336488112548 \\
0.15789473684210525 0.8954491076357266 \\
0.16842105263157894 0.8999575106092874 \\
0.17894736842105263 0.9036222205582479 \\
0.18947368421052632 0.9066243398036781 \\
0.2 0.9091440999824122 \\
    };
\addlegendentry{\textcolor{black}{DPFM: {0.67}}}    

\addplot [color=cPLOT3, smooth, line width=\pckLineWidth]
table[row sep=crcr]{%
0.0 0.13387466674560872 \\
0.010526315789473684 0.30691445179108456 \\
0.021052631578947368 0.53647557469513 \\
0.031578947368421054 0.6711887103599234 \\
0.042105263157894736 0.7441372476974756 \\
0.05263157894736842 0.7889208913716934 \\
0.06315789473684211 0.8188245414541472 \\
0.07368421052631578 0.8400239612301723 \\
0.08421052631578947 0.8551817030931929 \\
0.09473684210526316 0.8658946019318742 \\
0.10526315789473684 0.874153912593751 \\
0.11578947368421053 0.8804071667760827 \\
0.12631578947368421 0.8854623594497972 \\
0.1368421052631579 0.8897705050490979 \\
0.14736842105263157 0.8934534993670126 \\
0.15789473684210525 0.8964469117700836 \\
0.16842105263157894 0.8988787328409904 \\
0.17894736842105263 0.9011242274854117 \\
0.18947368421052632 0.9031198357541257 \\
0.2 0.9050370824416075 \\
    };
\addlegendentry{\textcolor{black}{URSSM: {0.79}}}  

\addplot [color=cPLOT1, smooth, line width=\pckLineWidth]
table[row sep=crcr]{%
0.0 0.13632564286970558 \\
0.010526315789473684 0.3398802983312466 \\
0.021052631578947368 0.572899779020341 \\
0.031578947368421054 0.6989487358535579 \\
0.042105263157894736 0.7650745915184908 \\
0.05263157894736842 0.8058304499173721 \\
0.06315789473684211 0.8332709342964262 \\
0.07368421052631578 0.8518443704169358 \\
0.08421052631578947 0.8643787406771486 \\
0.09473684210526316 0.8734286326252348 \\
0.10526315789473684 0.8802870123515266 \\
0.11578947368421053 0.885635625612744 \\
0.12631578947368421 0.89017276116609 \\
0.1368421052631579 0.8937782645870084 \\
0.14736842105263157 0.8968065043595159 \\
0.15789473684210525 0.8993506436968556 \\
0.16842105263157894 0.901604845183636 \\
0.17894736842105263 0.9036814270862901 \\
0.18947368421052632 0.9056570096175781 \\
0.2 0.9075185325139614 \\
    };
\addlegendentry{\textcolor{black}{Ours: \textbf{0.80}}}  

\end{axis}
\end{tikzpicture}  
    \end{tabular}
    } \\
    \end{tabular}
    \vspace{-4mm}
    \caption{Results on the SHREC'16 partial dataset. Our method substantially outperforms state-of-the-art methods, even in comparison to the supervised approaches. Meanwhile, our method demonstrates better cross-dataset generalisation ability.}
    \vspace{-3mm}
    \label{fig:shrec16}
\end{figure}

\noindent\textbf{Results.}~\cref{fig:shrec16-plot} and~\cref{fig:shrec16} provide the qualitative and quantitative results of our method in comparison to previous axiomatic, supervised and unsupervised methods that are suitable for partial shape matching. Similar to prior works~\cite{cao2022unsupervised,cao2023unsupervised}, we pre-train the feature extractor on complete shapes due to the limited amount of training shapes~\cite{ehm2024partial} and use vertex position as input features~\cite{attaiki2021dpfm}. Compared to existing methods, our approach is more robust to partiality and enables spatially smoother pointwise correspondences.

\section{Ablation Study}
\label{sec:ablation}
In this section we conduct ablative experiments to analyse the effectiveness of our regularisation. For all ablation experiments, we choose the most challenging TOPKIDS dataset for evaluation. We study a total of six different ablative settings and explain these in detail in the following. 

\noindent\textbf{(a) Removal of our regularisation.} We remove our regularisation $L_{\mathrm{diff}}$ defined in~\cref{eq:l_diff} for training. In this case, no spatial smoothness regularisation is taken into consideration.

\noindent\textbf{(b) Fixed single diffusion time.} We replace our randomly sampled diffusion time defined in~\cref{eq:t_i} by a heuristically manually selected single diffusion time applied to the whole distribution $\mathbf{F}_{\mathcal{M}}$ as defined in~\cref{eq:e_diff}.

\noindent\textbf{(c) Pre-defined initial distributions.} In~\cref{sec:method}, we choose randomly sampled initial distributions that we diffuse. In this ablative setting, we consider a pre-defined initial distributions. To this end, we choose to use the first $h$ Laplacian eigenfunctions $\mathbf{\Phi}_{\mathcal{M}} \in \mathbb{R}^{n_\mathcal{M} \times h}$, which corresponds to the smoothest orthogonal basis w.r.t. the Dirichlet energy~\cite{bobenko2007discrete}.   

\noindent\textbf{(d) Comparison to heat kernel matching.} In~\cref{sec:theoretical} we conceptually relate our regularisation to heat kernel matching~\cite{vestner2017efficient}, which we now study empirically. To this end, we replace $L_{\mathrm{diff}}$ defined in~\cref{eq:l_diff} by the quadratic assignment formulation defined in~\cref{eq:qap}. Similarly, we consider a multiscale setting, i.e.
\begin{equation}
    L_{\mathrm{kernel}} = \sum_{i=1}^{h}\left\| \mathbf{D}_{\mathcal{M}}^{t_i} - \mathbf{\Pi}_\mathcal{NM}^{\top} \mathbf{D}_{\mathcal{N}}^{t_i} \mathbf{\Pi}_\mathcal{NM} \right\|_F^2,
\end{equation}
where $\mathbf{D}_{\bullet}^{t_i} = \mathbf{\Phi}_{\bullet}\exp{\left(-t_i\mathbf{\Lambda}_{\bullet}\right)}\mathbf{\Phi}_{\bullet}^{\top}$ is the heat kernel, and $t_i$ is the randomly sampled diffusion time defined in~\cref{eq:t_i}. We emphasise that heat kernel matching is much more computationally expensive, since the matrix size in $L_{\mathrm{kernel}}$ is $n_{\mathcal{M}} \times n_{\mathcal{N}}$, opposed to $n_{\mathcal{M}} \times h$ in our regularisation.

\noindent\textbf{(e) Comparison to Dirichlet energy.} Many prior works~\cite{ezuz2019reversible,magnet2022smooth,cao2023unsupervised} utilise the Dirichlet energy to prompt smooth pointwise correspondences. The Dirichlet energy is expressed as
\begin{equation}
        \label{eq:dirichlet}
        L_{\mathrm{D}} = \left\|\mathbf{\Pi}_{\mathcal{NM}}\mathbf{V}_{\mathcal{M}}\right\|^{2}_{\mathbf{L}_{\mathcal{N}}} = 
        \text{tr}( (\mathbf{\Pi}_{\mathcal{NM}}\mathbf{V}_{\mathcal{M}})^\top \mathbf{L}_{\mathcal{N}}(\mathbf{\Pi}_{\mathcal{NM}}\mathbf{V}_{\mathcal{M}})),
\end{equation}
where $\mathbf{V}_{\mathcal{M}}$ is the matrix of vertex locations of shape $\mathcal{M}$.
However, the optimal solution of $L_{\mathrm{D}}$ leads to a degenerate result, i.e. all vertices of shape $\mathcal{M}$ are matched to the same vertex of shape $\mathcal{N}$. In contrast, our regularisation encourages bijective/high-coverage correspondences.

\noindent \textbf{(f) Comparison to cycle-consistency.} We point out that if $t_i = 0$ for all distributions (i.e.\ no diffusion), our regularisation degrades to the commonly used cycle-consistency loss~\cite{zhou2016learning,groueix2019unsupervised}, i.e.
\begin{equation}
    L_{\mathrm{diff}} = \left\|\mathbf{F}_{\mathcal{M}} - \mathbf{\Pi}_{\mathcal{MN}}\left(\mathbf{\Pi}_{\mathcal{NM}} \mathbf{F}_{\mathcal{M}}\right)\right\|^{2}_{F},
\end{equation}
since $h_{0}\left(u\right) = u$ in~\cref{eq:explicit_diffusion}.

\noindent\textbf{Results.} We summarise the quantitative results in~\cref{tab:ablation}. Compared to the use of a fixed diffusion time (see column (a)), we notice that our random diffusion time sample strategy enables multiscale regularisation, and thus obtains better performance. Compared to the pre-defined $\mathbf{F}_{\mathcal{M}}$ (see column (b)), we observe random initial distributions is better. In comparison to kernel matching (KM), Dirichlet energy, and cycle-consistency (column (d), column (e), and column (f)), our regularisation demonstrates better matching performance and significantly improves the baseline (i.e.\ compared to column (a)).
\begin{table}[bht!]
    \centering
    \scriptsize
    \vspace{-3mm}
    \caption{Qualitative results of our ablation study on TOPKIDS dataset. Compared to different settings, our proposed regularisation obtains the best matching performance.}
    \label{tab:ablation}
    \setlength{\tabcolsep}{1pt}
    \begin{tabular}{@{}lccccccc@{}}
    \toprule
    \textbf{TOPKIDS} & {(a) w/o $L_{\mathrm{diff}}$} & (b) Fixed $t$  & (c) Pre-defined $\mathbf{F}_{\mathcal{M}}$ & (d) KM & (e) Dirichlet & (f) Cycle-con. & Ours \\ \midrule
    \textbf{Initial $\mathbf{F}_{\mathcal{M}}$} & n/a & Rand. & $\mathbf{\Phi}_{\mathcal{M}}$ & Dirac & n/a & Rand. & Rand. \\
    \textbf{Geo. error} & 9.2 & 7.8 & 10.7 & 8.9 & 9.5 & 9.8 & \textbf{5.4} \\ 
    \bottomrule
\end{tabular}
\vspace{-3mm}   
\end{table}

\section{Discussion and Limitations}
\label{sec:limitation}
For the first time, we propose a simple yet efficient regularisation via synchronous diffusion for unsupervised smooth non-rigid 3D shape matching. Yet, there are some limitations that give rise to further investigations.

One limitation of our work is its scope: although conceptually our synchronous diffusion idea can be applied to other settings (including image keypoint matching~\cite{sarlin2020superglue}, graph matching~\cite{fey2019deep} and point cloud matching~\cite{cao2023self}), in this work we specifically focus on non-rigid 3D shape matching based on deep functional maps. As such,  we cannot handle partial-to-partial shape matching~\cite{attaiki2021dpfm}, which is a limitation that we inherit from the functional map framework~\cite{rodola2017partial}.
Analysing and extending our regularisation to the challenging and highly relevant case of  partial-to-partial shape matching is an interesting direction for future work.

Another limitation of our work is that -- from a theoretical point of view -- the diffusion process is inconsistent for pairs of shapes with severe non-isometries and topological differences. Yet, experimentally we demonstrate that our regularisation works well in such settings. However, this requires a careful choice of the maximum diffusion time $T$ to adequately balance local and global information exchange (see supplementary for more details).

\section{Conclusion}
\label{sec:conclusion}
In this work we propose the first unsupervised regularisation based on a synchronous diffusion process for smooth non-rigid 3D shape matching. We demonstrate that our simple regularisation can  easily be integrated into a recent state-of-the-art shape matching method and that this leads to a notable performance improvement in different challenging cases, including non-isometric, topologically noisy and partial shapes. Our work is the first attempt to design a diffusion-based smoothness regularisation for unsupervised learning of matching problems and we hope that it will inspire follow-up works across different domains and variants of matching problems. %

\clearpage
\bibliographystyle{splncs04}
\bibliography{main}

\clearpage
\setcounter{page}{1}
\title{Supplementary Material: Synchronous Diffusion for Unsupervised Smooth Non-Rigid 3D Shape Matching} 

\titlerunning{Synchronous Diffusion for Unsupervised 3D Shape Matching}

\author{Dongliang Cao\inst{1}\orcidlink{0000-0002-6505-6465} \and
Zorah Lähner\inst{1}\orcidlink{0000-0003-0599-094X} \and
Florian Bernard\inst{1}\orcidlink{0009-0008-1137-0003}}

\authorrunning{Dongliang Cao et al.}

\institute{University of Bonn}

\maketitle

In this supplementary document, we first provide a background on functional maps~\cite{ovsjanikov2012functional,roufosse2019unsupervised} (Section~\ref{sec:deep_functional_map}) and the deep shape matching block~\cite{cao2023unsupervised} (Section~\ref{sec:deep_matching_block}) used in our framework. Afterwards, we provide the implementation details of our method in Section~\ref{sec:implementation}. Next, we conduct an ablative experiment to investigate the influence of the choice of diffusion time (Section~\ref{sec:diff_time}). Eventually, we show more qualitative results of our method  in Section~\ref{sec:qual_exp}.

\section{Deep Functional Maps in a Nutshell}
\label{sec:deep_functional_map}
In this section, we provide a brief introduction to the standard pipeline of the deep functional maps method~\cite{roufosse2019unsupervised}. We consider a pair of 3D shapes $\mathcal{M}$ and $\mathcal{N}$ represented as triangle meshes with $n_{\mathcal{M}}$ and $n_{\mathcal{N}}$ (w.l.o.g.\ $n_{\mathcal{M}} \leq n_{\mathcal{N}}$) vertices, respectively. Here we summarise the main steps of its pipeline:
\begin{enumerate}
    \item Compute the Laplacian matrices $\mathbf{L}_{\mathcal{M}} \in \mathbb{R}^{n_{\mathcal{M}} \times n_{\mathcal{M}}}, \mathbf{L}_{\mathcal{N}} \in \mathbb{R}^{n_{\mathcal{N}} \times n_{\mathcal{N}}}$~\cite{pinkall1993computing} and the corresponding first $k$ eigenfunctions $\mathbf{\Phi}_{\mathcal{M}} \in \mathbb{R}^{n_{\mathcal{M}} \times k}, \mathbf{\Phi}_{\mathcal{N}} \in \mathbb{R}^{n_{\mathcal{N}} \times k}$ and the diagonal eigenvalue matrix $\mathbf{\Lambda}_{\mathcal{M}} \in \mathbb{R}^{k \times k}, \mathbf{\Lambda}_{\mathcal{N}} \in \mathbb{R}^{k \times k}$, respectively.
    \item Compute pointwise features $\mathbf{E}_{\mathcal{M}} \in \mathbb{R}^{n_{\mathcal{M}} \times d}, \mathbf{E}_{\mathcal{N}} \in \mathbb{R}^{n_{\mathcal{N}} \times d}$ defined on each shape via a learnable feature extractor $\mathcal{F}_{\theta}$ with weights $\theta$.
    \item Compute the functional map $\mathbf{C}_{\mathcal{MN}} \in \mathbb{R}^{k \times k}$ associated with the Laplacian eigenfunctions by solving the optimisation problem
    \begin{equation}
    \label{eq:fmap}
        \mathbf{C}_{\mathcal{MN}}=\arg\min_{\mathbf{C}}~ E_{\mathrm{data}}\left(\mathbf{C}\right)+\lambda E_{\mathrm{reg}}\left(\mathbf{C}\right).
    \end{equation}
    Here, $E_{\mathrm{data}}\left(\mathbf{C}\right)=\left\|\mathbf{C}\mathbf{\Phi}_{\mathcal{M}}^{\dagger}\mathbf{E}_{\mathcal{M}}-\mathbf{\Phi}_{\mathcal{N}}^{\dagger}\mathbf{E}_{\mathcal{N}}\right\|^{2}_{F}$ enforces descriptor preservation, while the regularisation term $E_{\mathrm{reg}}$ imposes certain structural properties (e.g.\ Laplacian commutativity $E_{\mathrm{reg}}\left(\mathbf{C}\right)=\left\|\mathbf{C}\mathbf{\Lambda}_{\mathcal{M}}-\mathbf{\Lambda}_{\mathcal{N}}\mathbf{C}\right\|^{2}_{F}$~\cite{ovsjanikov2012functional}).
    \item During training, structural regularisation (e.g.\ orthogonality, bijectivity~\cite{roufosse2019unsupervised}) is imposed on the functional maps, i.e.
    \begin{equation}
        \label{eq:l_struct}
        L_{\mathrm{struct}} = \lambda_{\mathrm{bij}}L_{\mathrm{bij}} + \lambda_{\mathrm{orth}}L_{\mathrm{orth}}.
    \end{equation}
    \item During inference, the point-wise map $\mathbf{\Pi}_{\mathcal{MN}}$ is obtained based on the map relationship $\mathbf{C}_{\mathcal{NM}} = \mathbf{\Phi}_{\mathcal{M}}^{\dagger}\mathbf{\Pi}_{\mathcal{MN}}\mathbf{\Phi}_{\mathcal{N}}$, e.g.\ either by nearest neighbour search in the spectral domain or by other post-processing techniques~\cite{vestner2017product,melzi2019zoomout,eisenberger2020smooth,ezuz2019reversible}.
\end{enumerate}  

\section{Deep Shape Matching Block}
\label{sec:deep_matching_block}
In this section, we provide a brief introduction about our deep shape matching block in~\cref{fig:sync_diffusion} based on the shape matching framework proposed in~\cite{cao2023unsupervised}. 
\subsection{Pointwise Correspondences Based on Feature Matching}
In theory, the pointwise map $\mathbf{\Pi}_{\mathcal{MN}}$ should be a permutation matrix, i.e.
    \begin{equation}
        \label{eq:permutation_mat}
        \left\{\mathbf{\Pi} \in\{0,1\}^{n_{\mathcal{M}} \times n_{\mathcal{N}}}: \mathbf{\Pi} \mathbf{1}_{n_{\mathcal{N}}} = \mathbf{1}_{n_{\mathcal{M}}}, \mathbf{1}_{n_{\mathcal{M}}}^{\top} \mathbf{\Pi} \leq \mathbf{1}_{n_{\mathcal{N}}}^{\top}\right\},
    \end{equation}
where $\mathbf{\Pi}(i,j)$ indicates whether the $i$-th vertex in shape $\mathcal{M}$ corresponds to the $j$-th vertex in shape $\mathcal{N}$. In our work the pointwise correspondences $\mathbf{\Pi}_{\mathcal{MN}}$ between shapes $\mathcal{M}$ and $\mathcal{N}$ are obtained based on pairwise similarity of the learned features $\mathbf{E}_{\mathcal{M}}, \mathbf{E}_{\mathcal{N}}$. Following prior works~\cite{cao2023unsupervised,eisenberger2021neuromorph}, we use the softmax operator to produce a soft correspondence matrix, i.e.
    $\mathbf{\Pi}_{\mathcal{MN}} = \mathrm{Softmax}\left( {\mathbf{E}_{\mathcal{M}}\mathbf{E}_{\mathcal{N}}^{T}/\tau} \right),$
where $\tau$ is the temperature factor to determine the softness of the pointwise map. The softmax operator is applied in each row to ensure that correspondences are non-negative and $\mathbf{\Pi}_{\mathcal{MN}} \mathbf{1}_{n_{\mathcal{N}}} = \mathbf{1}_{n_{\mathcal{M}}}$. In this way, $\mathbf{\Pi}_{\mathcal{MN}}$ can be interpreted as a soft assignment of vertices $\mathbf{V}_{\mathcal{N}}$ in shape $\mathcal{N}$ to vertices $\mathbf{V}_{\mathcal{M}}$ in shape $\mathcal{M}$. Similarly, we can obtain $\mathbf{\Pi}_{\mathcal{NM}}$ with the roles of $\mathcal{M}$ and $\mathcal{N}$ swapped. 

\subsection{Spectral Regularisation on Pointwise Correspondences}
The predicted pointwise soft correspondences are regularised in the spectral domain based on the relationship to the associated functional map. Specifically, the predicted pointwise maps $\mathbf{\Pi}_{\mathcal{MN}}, \mathbf{\Pi}_{\mathcal{NM}}$ are regularised by the coupling relationship between pointwise maps and functional maps~\cite{ren2021discrete}, i.e.
\begin{equation}
    \label{eq:l_couple}
    L_{\mathrm{couple}} = \left\|\mathbf{C}_{\mathcal{MN}} - \mathbf{\Phi}_{\mathcal{N}}^{\dagger}\mathbf{\Pi}_{\mathcal{NM}}\mathbf{\Phi}_{\mathcal{M}}\right\|^{2}_{F}
    + \left\|\mathbf{C}_{\mathcal{NM}} - \mathbf{\Phi}_{\mathcal{M}}^{\dagger}\mathbf{\Pi}_{\mathcal{MN}}\mathbf{\Phi}_{\mathcal{N}}\right\|^{2}_{F},
\end{equation}
where $\mathbf{C}_{\mathcal{MN}}, \mathbf{C}_{\mathcal{NM}}$ are the functional maps computed by solving the optimisation problem in~\cref{eq:fmap}. To this end, our total loss is a linear combination of our synchronous diffusion regularisation and spectral regularisation related to functional maps, i.e.
\begin{equation}
    \label{eq:total}
    L_{\mathrm{total}} = L_{\mathrm{diff}} + \lambda_{\mathrm{couple}}L_{\mathrm{couple}} + \lambda_{\mathrm{struct}}L_{\mathrm{struct}}.
\end{equation}

\section{Implementation Details} \label{sec:implementation}
As we described in~\cref{sec:deep_matching_block}, our deep shape matching block is based on the framework proposed in~\cite{cao2023unsupervised}. To this end, we use the official implementation of~\cite{cao2023unsupervised} to build our deep shape matching block. In the context of our regularisation based on synchronous diffusion, the dimension of the random distribution $\mathbf{F}_{\bullet}$ is 128 (i.e.\ $h=128$). Regarding to the maximum diffusion time in~\cref{eq:t_i}, we heuristically choice $T=1.0^{-2}$ for near-isometric, topological noise and partial shape datasets and $T=1.0^{-4}$ for non-isometric shape datasets, respectively. Since for non-isometric shape pairs, the longer diffusion time causes more inconsistent diffusion processes on both shapes. For the total loss we set $\lambda_{\mathrm{couple}} = 1, \lambda_{\mathrm{struct}} = 1$ in~\cref{eq:total} based on empirical experiments. In the context of input features, we follow previous works~\cite{cao2022unsupervised,cao2023unsupervised,attaiki2021dpfm} by using WKS~\cite{aubry2011wave} as input features except for partial shape matching on SHREC'16 dataset, where vertex positions are used. When we evaluate our method on SHREC'16 dataset, we pre-train the feature extractor on complete shapes by using a combination of complete shape datasets (DT4D-H~\cite{magnet2022smooth},
SMAL~\cite{zuffi20173d}, FAUST~\cite{bogo2014faust} and SCAPE~\cite{anguelov2005scape}). 

\section{Influence of Diffusion Time} \label{sec:diff_time}
Here, we conduct an ablative experiment to investigate the influence of the choice of the maximum diffusion time on the challenging TOPKIDS dataset and choose the optimal value for our final results. See Table~\ref{tab:ablation_time}. While large values for $T$ are likely to be affected by  topological noise, a too small value is not able to provide any benefit due to only small neighbourhood being covered. 

\begin{table}[bht!]
    \centering
    \scriptsize
    \vspace{-3mm}
    \caption{Qualitative results on TOPKIDS dataset with different maximum diffusion time. A careful choice of the maximum diffusion time is important for our regularisation.}
    \label{tab:ablation_time}
    \setlength{\tabcolsep}{8pt}
    \begin{tabular}{@{}lcccccc@{}}
    \toprule
    \textbf{Diff. Time $T$} & {$1.0$} & $1.0^{-1}$  & $1.0^{-2}$ (Ours) & $5.0^{-3}$ & $1.0^{-3}$ & $1.0^{-4}$ \\ \midrule
    \textbf{Geo. error ($\times 100$)} & 23.7 & 10.5 & \textbf{5.4} & 6.9 & 8.5 & 10.3 \\ 
    \bottomrule
\end{tabular}
\vspace{-8mm}   
\end{table}

\section{More Qualitative Results} \label{sec:qual_exp}

In the next figures, we provide additional qualitative results of our method on TOPKIDS, SHREC'19, SMAL and DTH4D-H corresponding to the quantitative results reported in the main text. 

\begin{figure}[bht!]
    \centering
\def\rowOnecolumnOne{kid00-kid04}
\def\rowOnecolumnTwo{kid00-kid04}
\def\rowOnecolumnThree{kid00-kid18}
\def\rowOnecolumnFour{kid00-kid07}
\def\rowOnecolumnFive{kid00-kid24}
\def\rowOnecolumnSix{kid00-kid25}

\def\pathShrecNT{figs/ours/topkids/}
\def\hspaceCols{0.13cm}
\def\wspaceRows{0cm}
\def\height{3.1cm}
\def\width{3.0cm}
\def\heightT{\height}
\def\widthT{\width}
\begin{tabular}{cccccc}%
    \setlength{\tabcolsep}{0pt} 
    {\scriptsize Source} & & & & & \\
    \vspace{\wspaceRows}
    \hspace{\hspaceCols}
    \includegraphics[height=\heightT, width=\widthT]{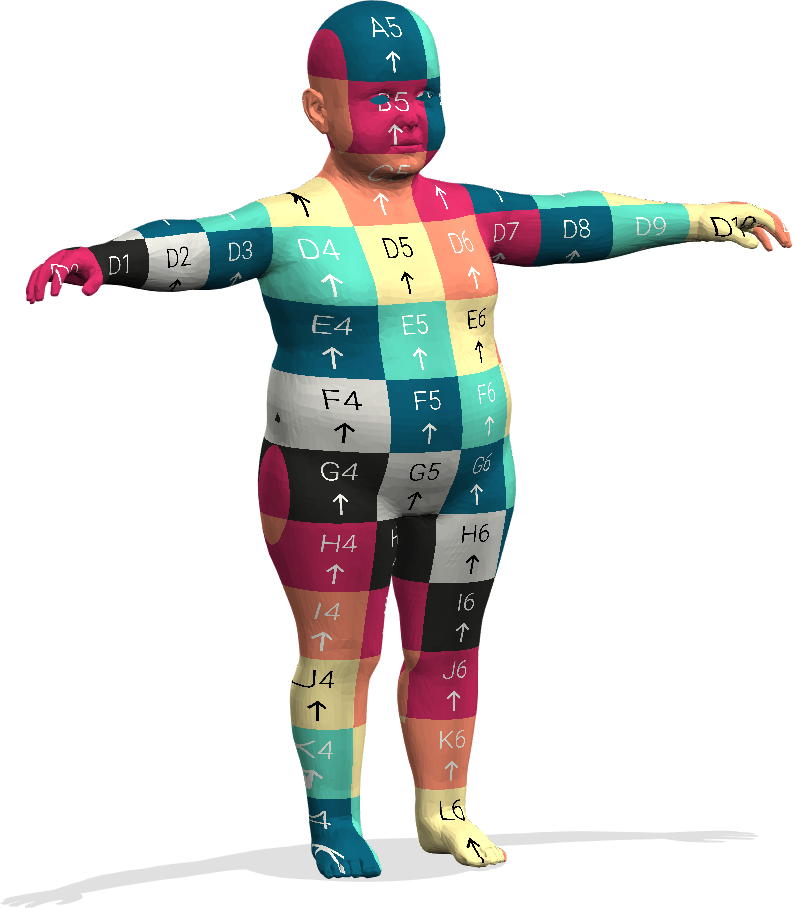}&
    \hspace{\hspaceCols}
    \includegraphics[height=\heightT, width=\widthT]{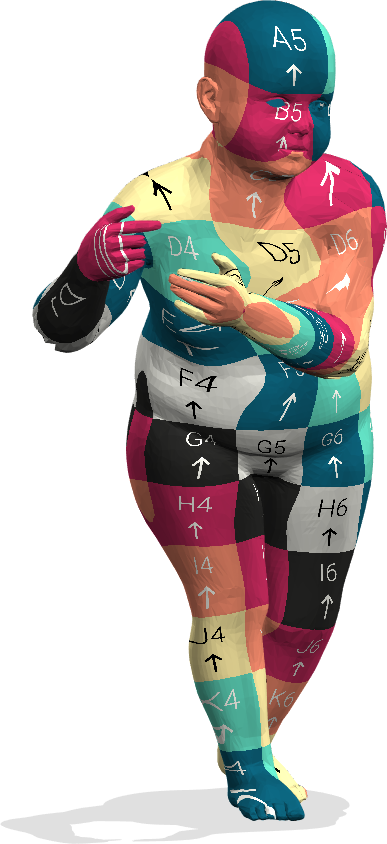}&
    \hspace{\hspaceCols}
    \includegraphics[height=\heightT, width=\widthT]{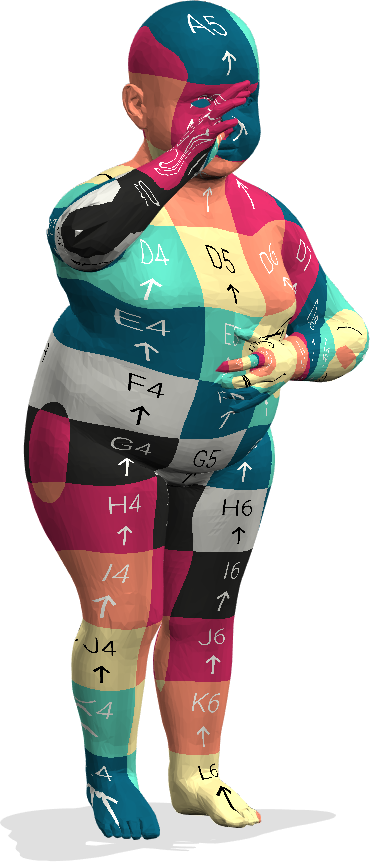}&
    \hspace{\hspaceCols}
    \includegraphics[height=\heightT, width=\widthT]{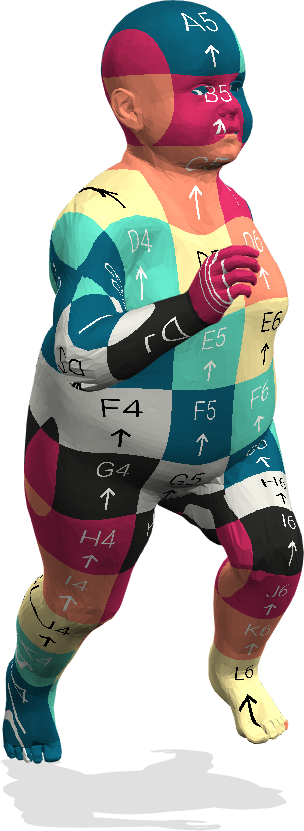}&
    \hspace{\hspaceCols}
    \includegraphics[height=2.8cm, width=\widthT]{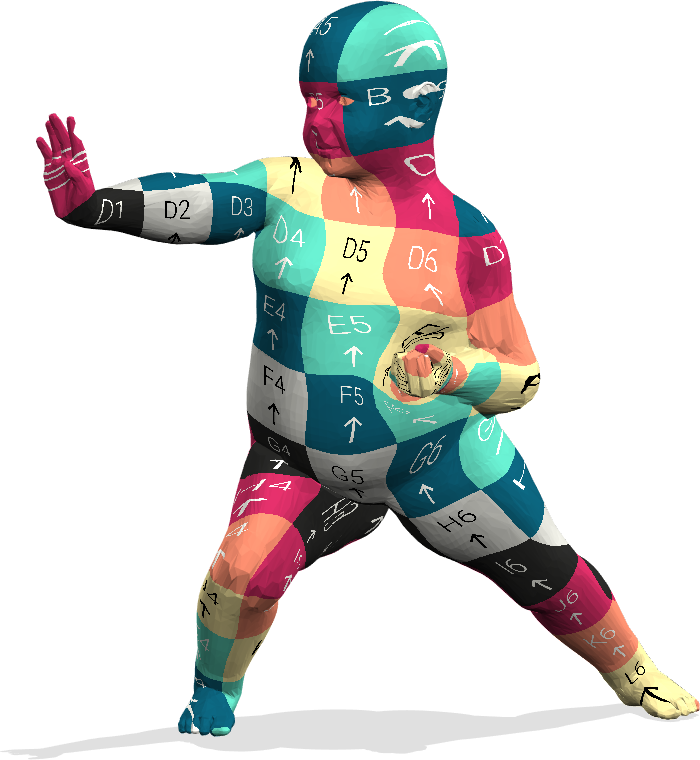}
    &
    \hspace{\hspaceCols}
    \includegraphics[height=\heightT, width=\widthT]{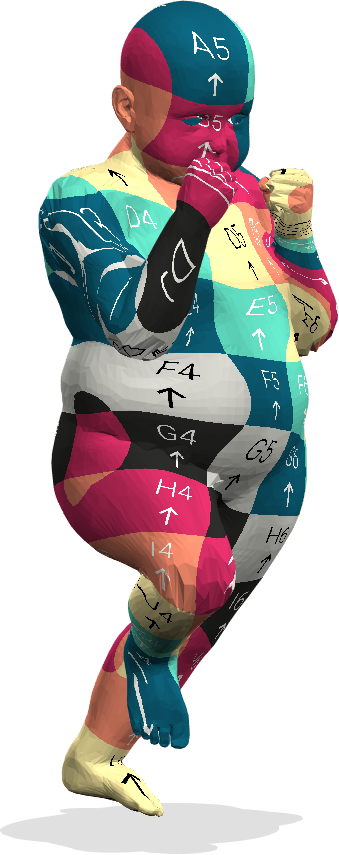}
    \\
\end{tabular}
    \caption{\textbf{Qualitative results of our method on TOPKIDS.} Our method obtains accurate correspondences for shapes with topological noise.}
    \label{fig:topkids_sup}
\end{figure}

\begin{figure}[bht!]
    \centering
\def\rowOnecolumnOne{2-11}
\def\rowOnecolumnTwo{2-18}
\def\rowOnecolumnThree{2-20}
\def\rowOnecolumnFour{2-39}
\def\rowOnecolumnFive{2-26}

\def\rowTwocolumnOne{2-34}
\def\rowTwocolumnTwo{2-36}
\def\rowTwocolumnThree{2-23}
\def\rowTwocolumnFour{2-44}
\def\rowTwocolumnFive{2-11}
\def\pathShrecNT{figs/ours/shrec19/}
\def\hspaceCols{0.0cm}
\def\wspaceRows{0cm}
\def\height{3.8cm}
\def\width{3.4cm}
\def\heightT{\height}
\def\widthT{\width}
\begin{tabular}{ccccc}%
    \setlength{\tabcolsep}{0pt} 
    {\scriptsize Source} & & & & \\
    \vspace{\wspaceRows}
    \hspace{\hspaceCols}
    \includegraphics[height=\heightT, width=\widthT]{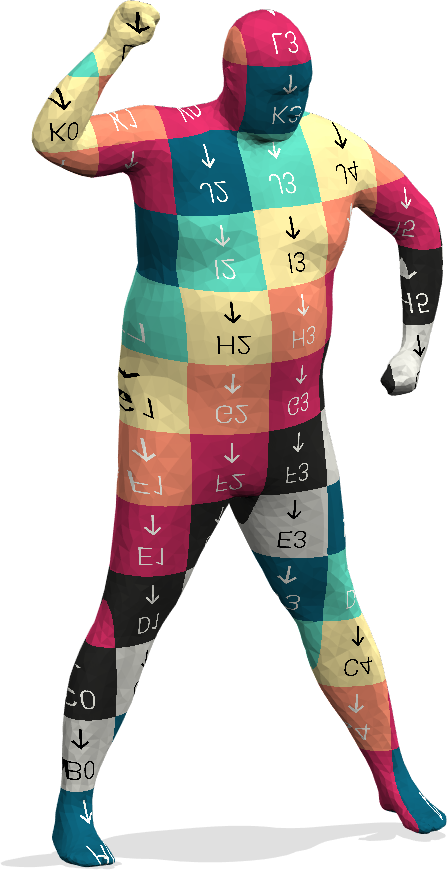}&
    \hspace{\hspaceCols}
    \includegraphics[height=\heightT, width=\widthT]{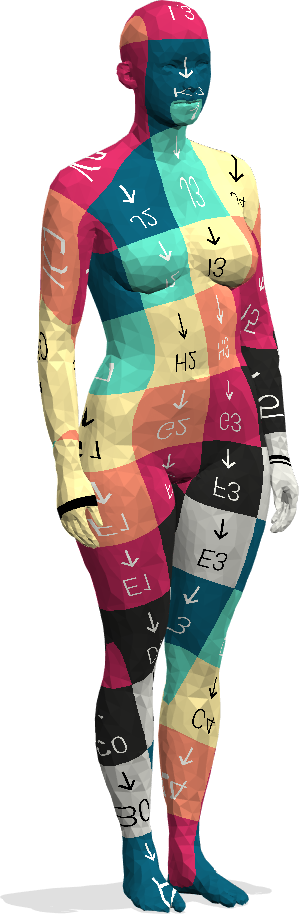}&
    \hspace{\hspaceCols}
    \includegraphics[height=\heightT, width=\widthT]{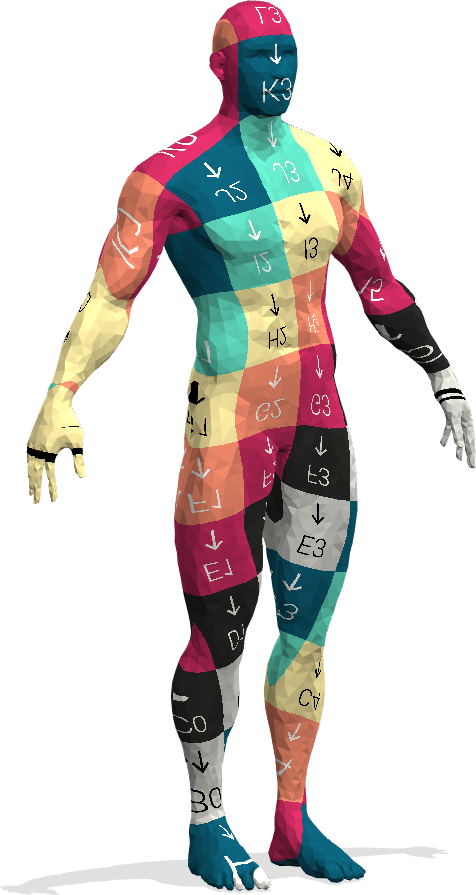}&
    \hspace{\hspaceCols}
    \includegraphics[height=3.2cm, width=\widthT]{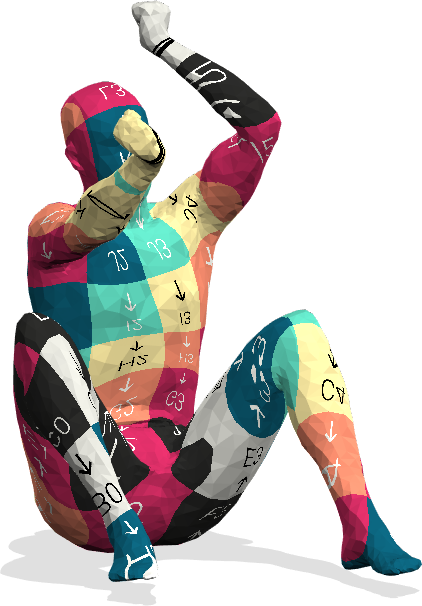}&
    \hspace{\hspaceCols}
    \includegraphics[height=\heightT, width=\widthT]{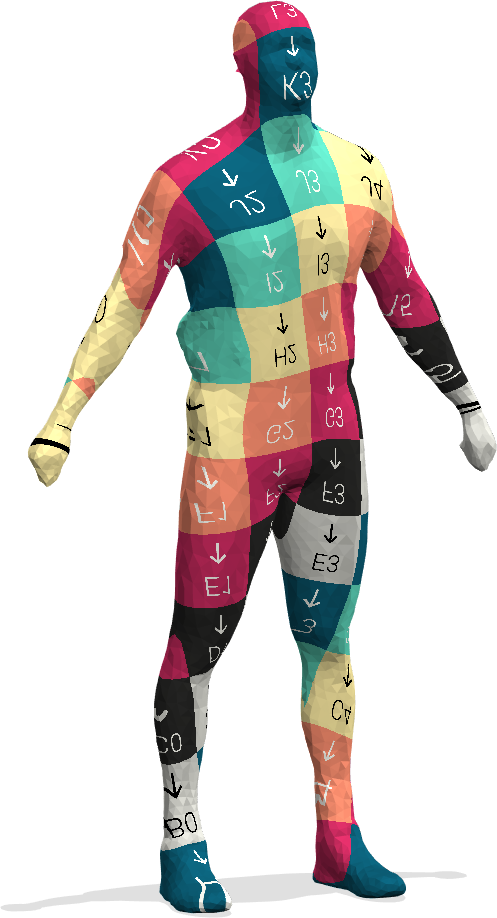}
    \\
    \hspace{\hspaceCols}
    \includegraphics[height=\heightT, width=\widthT]{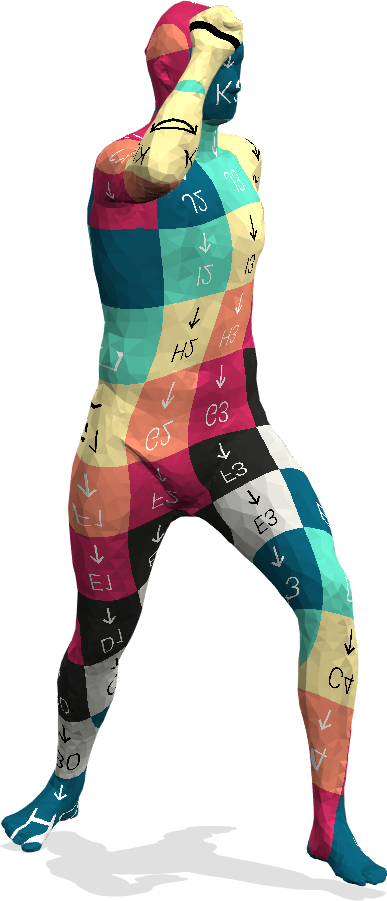}&
    \hspace{\hspaceCols}
    \includegraphics[height=4.0cm, width=\widthT]{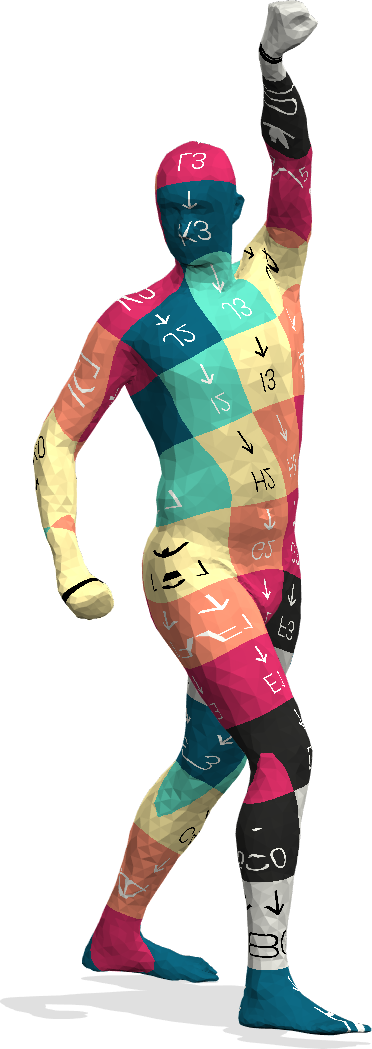}
    &
    \hspace{\hspaceCols}
    \includegraphics[height=\heightT, width=\widthT]{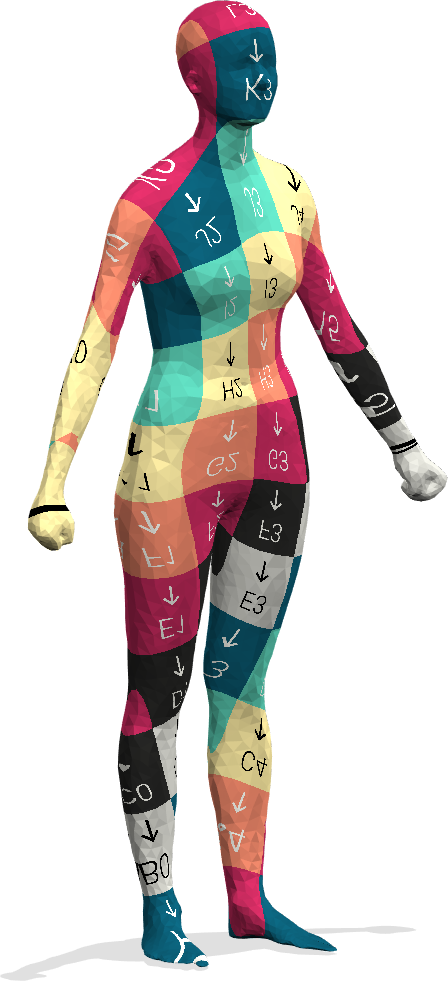}&
    \hspace{\hspaceCols}
    \includegraphics[height=\heightT, width=\widthT]{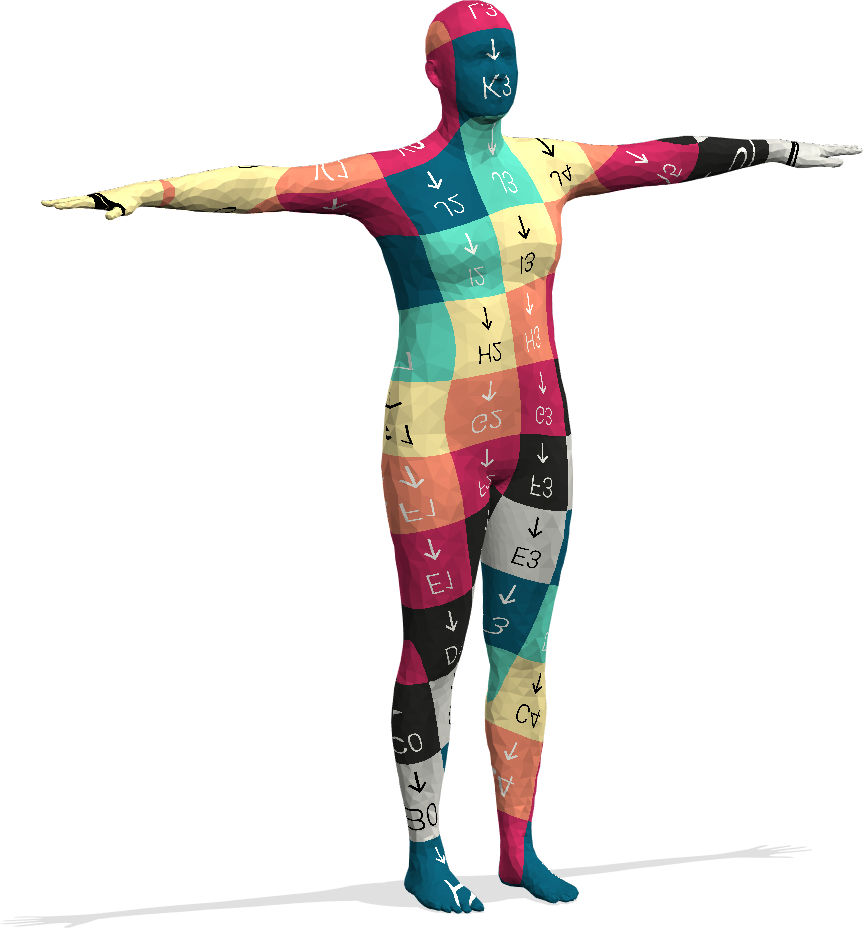}&
    \hspace{\hspaceCols}
    \includegraphics[height=\heightT, width=\widthT]{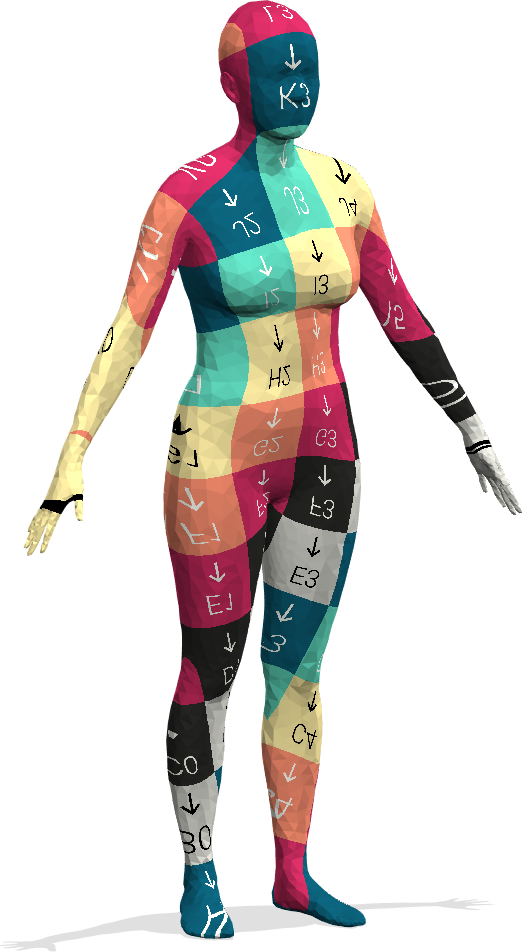}
\end{tabular}
    \caption{\textbf{Qualitative results of our method on SHREC'19.} Our method obtains accurate correspondences for human shapes with diverse poses and shapes.}
    \label{fig:shrec19}
\end{figure}

\begin{figure}[bht!]
    \centering
\def\rowOnecolumnOne{horse_01-cougar_01}
\def\rowOnecolumnTwo{horse_01-cougar_01}
\def\rowOnecolumnThree{horse_01-hippo_03}
\def\rowOnecolumnFour{horse_01-horse_08}
\def\pathShrecNT{figs/ours/smal/}
\def\hspaceCols{0.2cm}
\def\wspaceRows{0cm}
\def\height{2.8cm}
\def\width{2.7cm}
\def\heightT{\height}
\def\widthT{\width}
\begin{tabular}{cccc}%
    \setlength{\tabcolsep}{2pt} 
    {\scriptsize Source} & & &  \\
    \vspace{\wspaceRows}
    \hspace{\hspaceCols}
    \includegraphics[height=\heightT, width=\widthT]{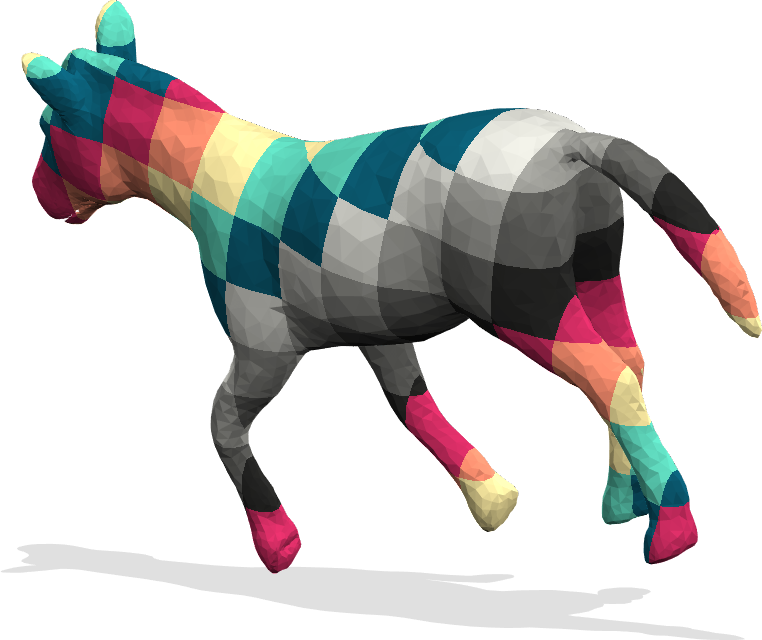}&
    \hspace{\hspaceCols}
    \includegraphics[height=\heightT, width=\widthT]{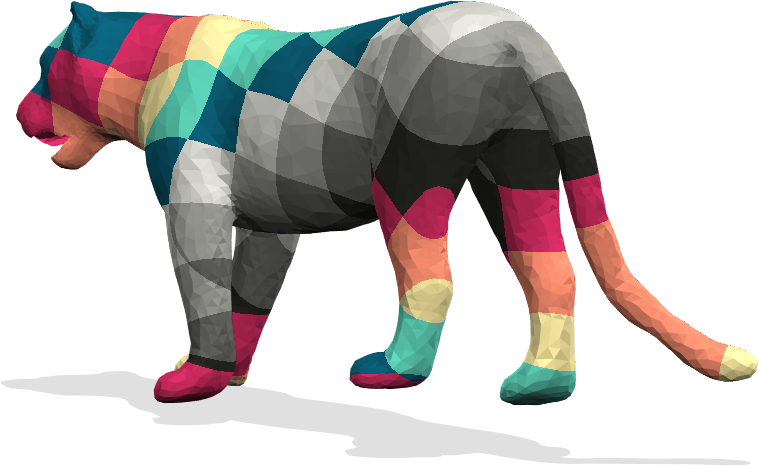}&
    \hspace{\hspaceCols}
    \includegraphics[height=\heightT, width=\widthT]{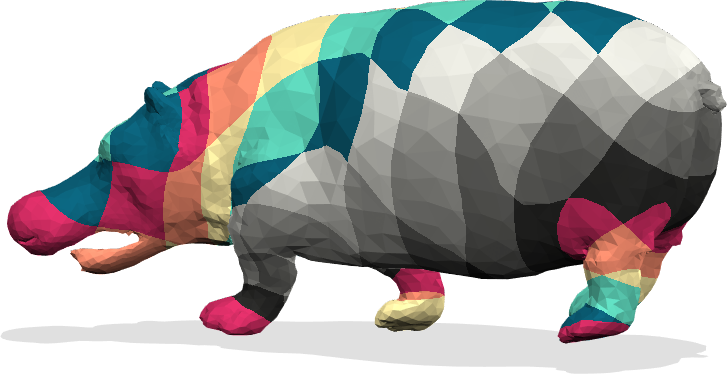}&
    \hspace{\hspaceCols}
    \includegraphics[height=\heightT, width=\widthT]{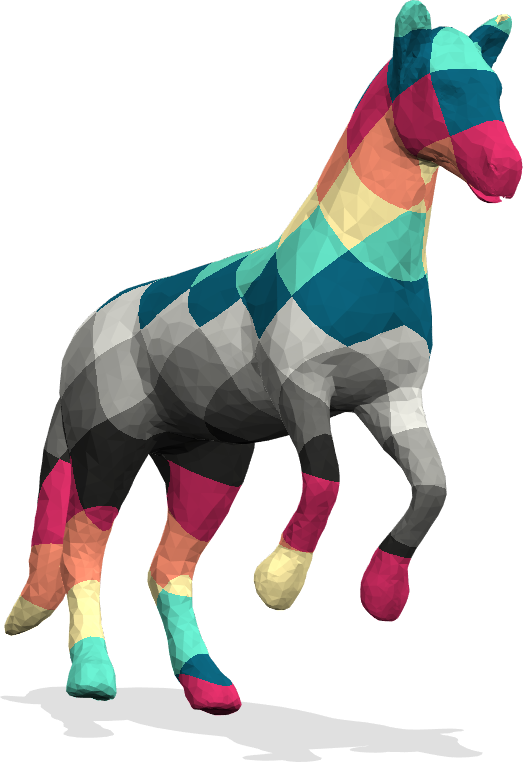}
    \\
\end{tabular}
    \caption{\textbf{Qualitative results of our method on SMAL.} Our method obtains accurate correspondences for shapes in different classes.}
    \label{fig:smal}
\end{figure}

\begin{figure}[bht!]
    \centering
\def\rowOnecolumnOne{Standing2HMagicAttack01036-Falling257}
\def\rowOnecolumnTwo{Standing2HMagicAttack01036-Falling257}
\def\rowOnecolumnThree{Standing2HMagicAttack01036-StandingCoverTurn015}
\def\rowOnecolumnFour{Standing2HMagicAttack01036-Floating029}
\def\rowOnecolumnFive{Standing2HMagicAttack01036-DancingRunningMan259}

\def\rowTwocolumnOne{Standing2HMagicAttack01036-Floating099}
\def\rowTwocolumnTwo{Standing2HMagicAttack01036-InvertedDoubleKickToKipUp189}
\def\rowTwocolumnThree{Standing2HMagicAttack01036-KettlebellSwing047}
\def\rowTwocolumnFour{Standing2HMagicAttack01036-GoalkeeperScoop061}
\def\rowTwocolumnFive{Standing2HMagicAttack01036-Strafing009}
\def\pathShrecNT{figs/ours/dt4d/}
\def\hspaceCols{0.0cm}
\def\wspaceRows{0cm}
\def\height{3.1cm}
\def\width{3.0cm}
\def\heightT{\height}
\def\widthT{\width}
\begin{tabular}{ccccc}%
    \setlength{\tabcolsep}{0pt} 
    {\scriptsize Source} & & & & \\
    \vspace{\wspaceRows}
    \hspace{\hspaceCols}
    \includegraphics[height=\heightT, width=\widthT]{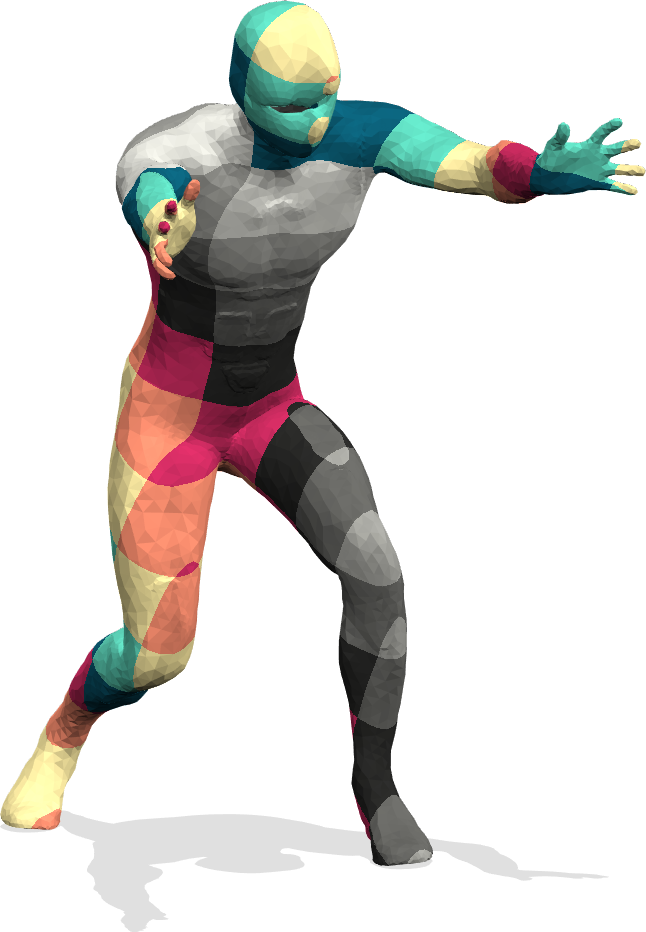}&
    \hspace{\hspaceCols}
    \includegraphics[height=\heightT, width=\widthT]{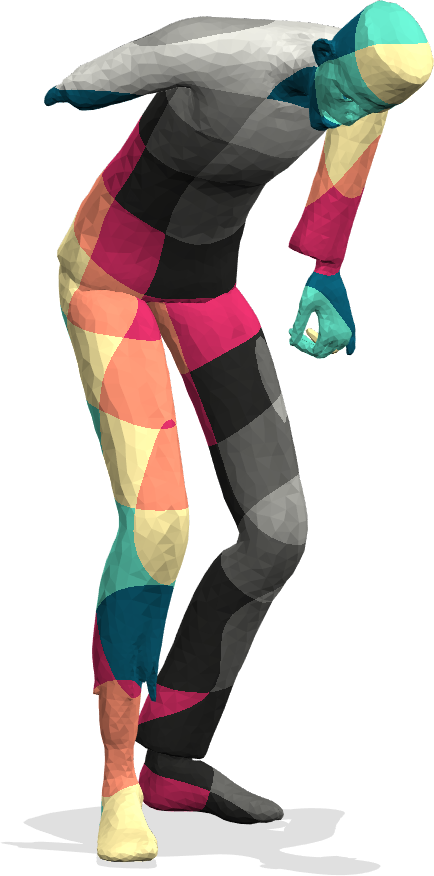}&
    \hspace{\hspaceCols}
    \includegraphics[height=\heightT, width=\widthT]{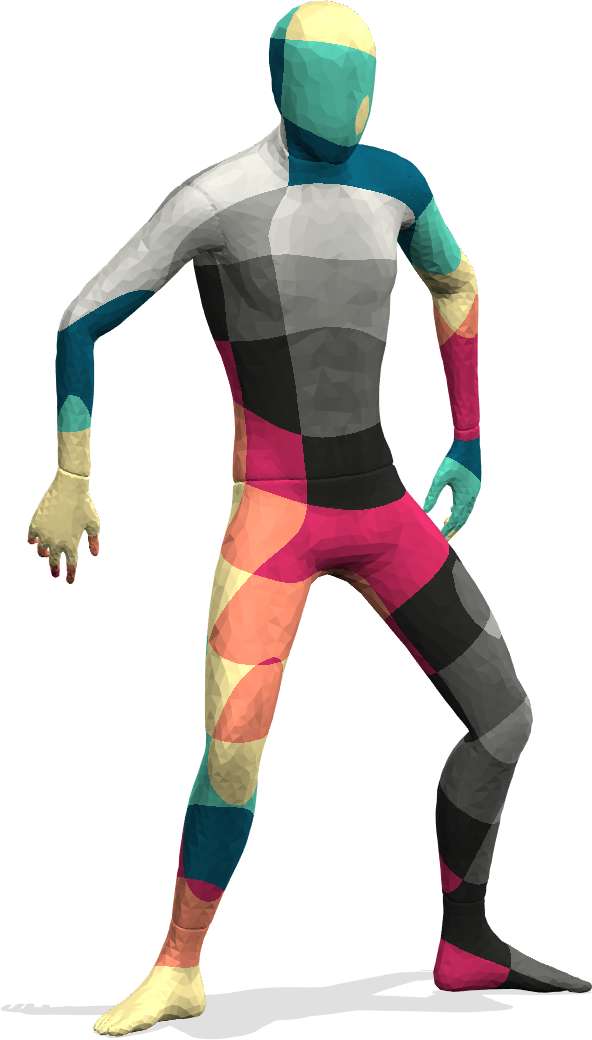}&
    \hspace{\hspaceCols}
    \includegraphics[height=\heightT, width=\widthT]{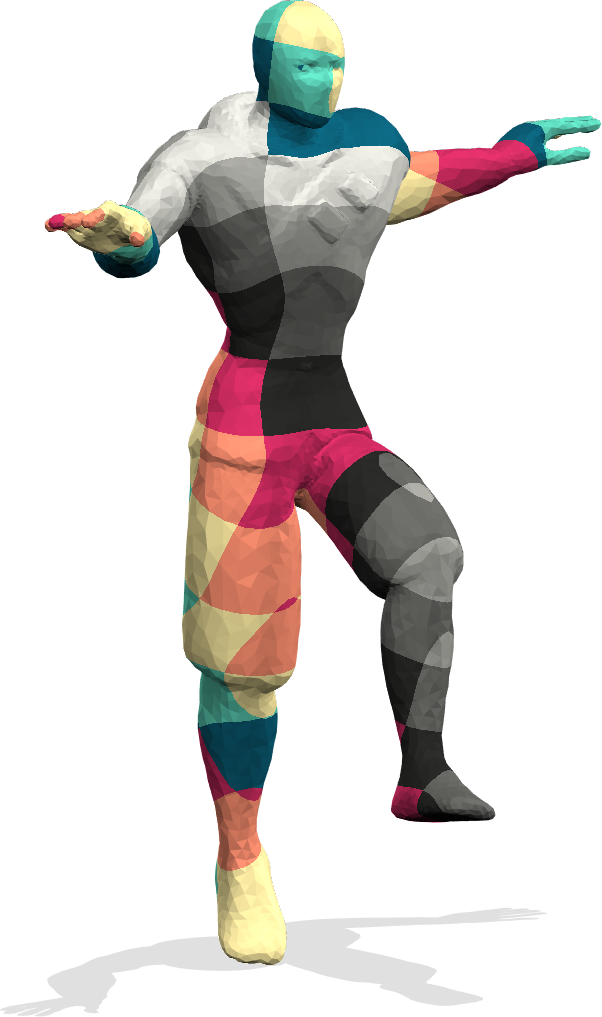}&
    \hspace{\hspaceCols}
    \includegraphics[height=\heightT, width=\widthT]{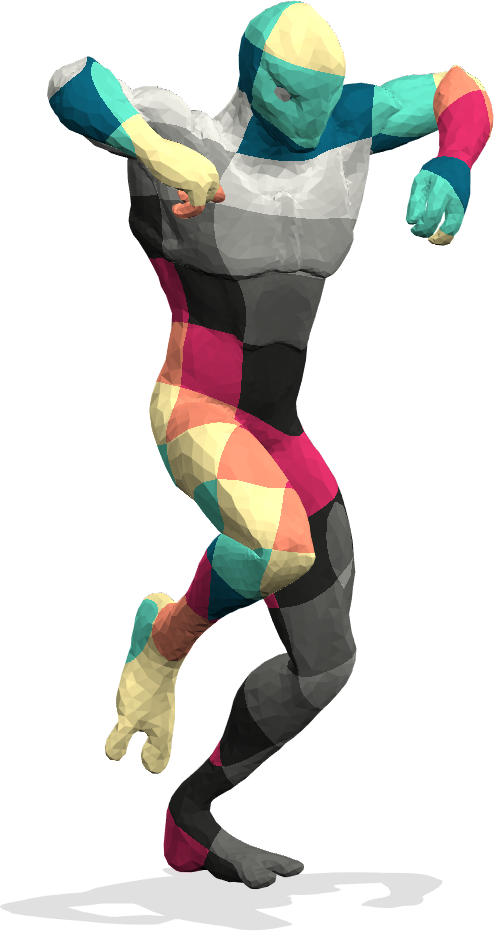}
    \\
    \hspace{\hspaceCols}
    \includegraphics[height=\heightT, width=\widthT]{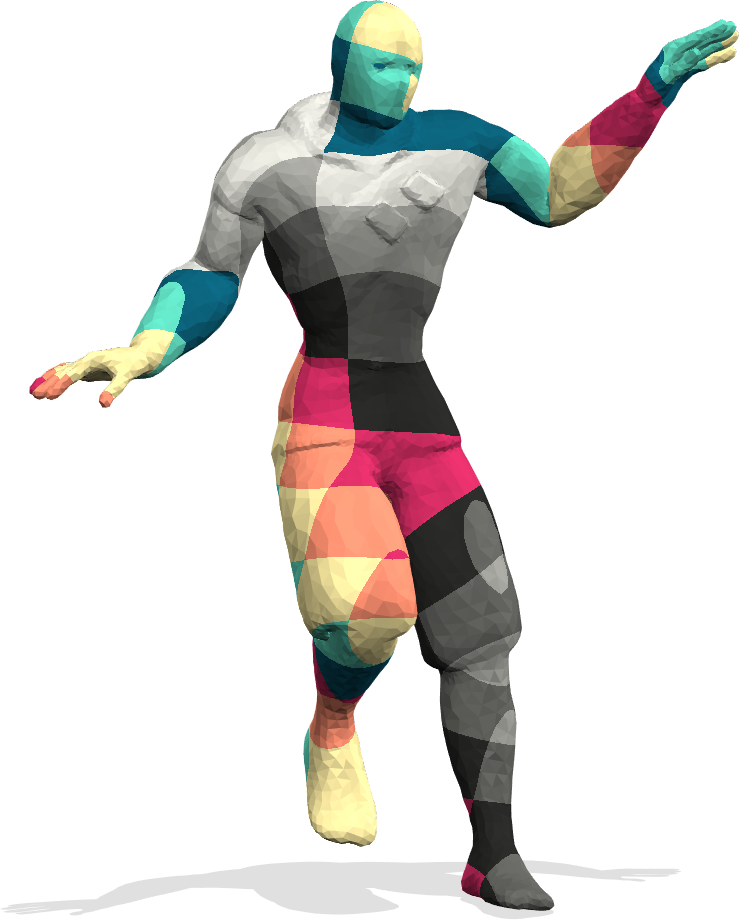}&
    \hspace{\hspaceCols}
    \includegraphics[height=\heightT, width=\widthT]{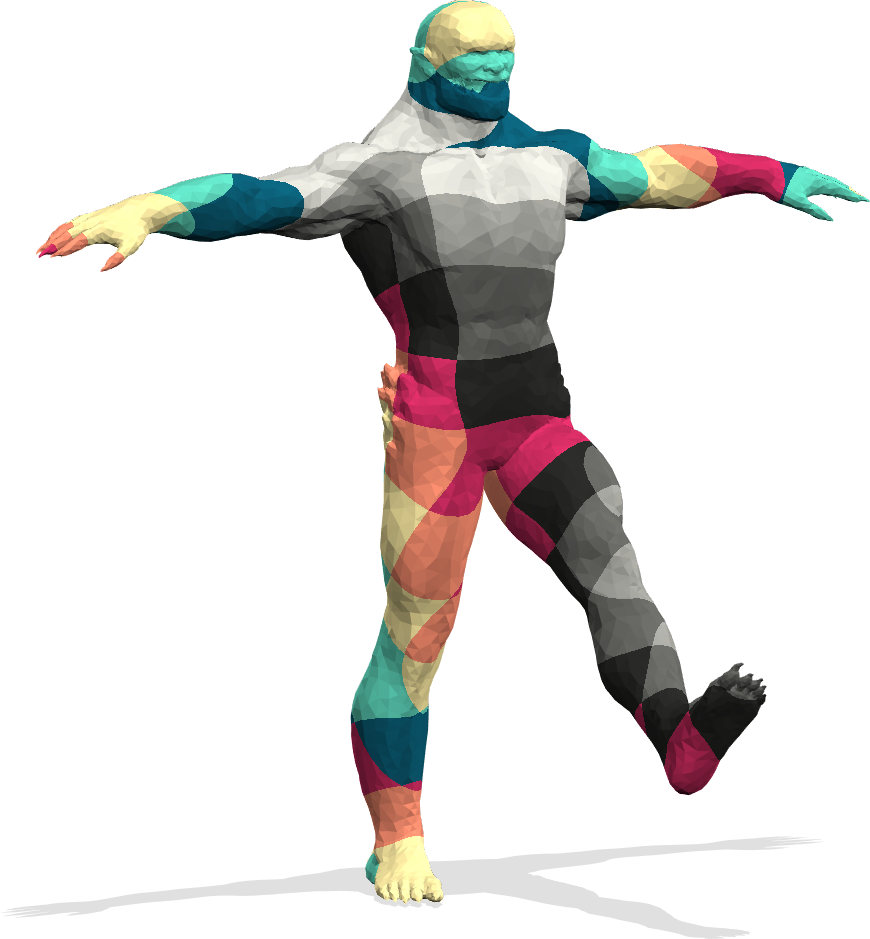}
    &
    \hspace{\hspaceCols}
    \includegraphics[height=\heightT, width=\widthT]{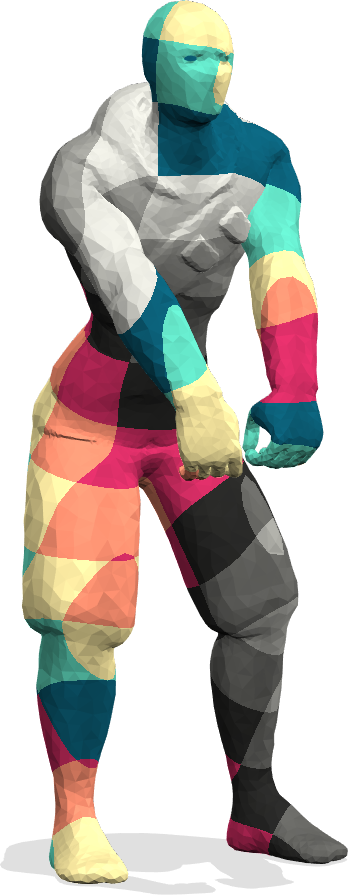}&
    \hspace{\hspaceCols}
    \includegraphics[height=\heightT, width=\widthT]{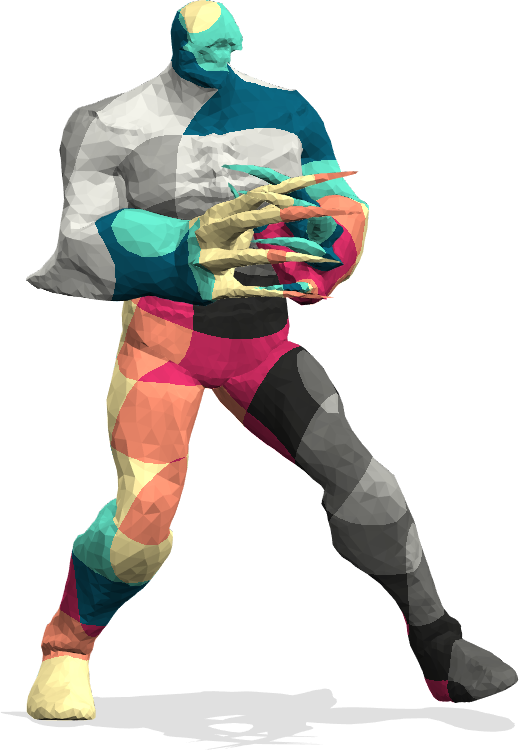}&
    \hspace{\hspaceCols}
    \includegraphics[height=\heightT, width=\widthT]{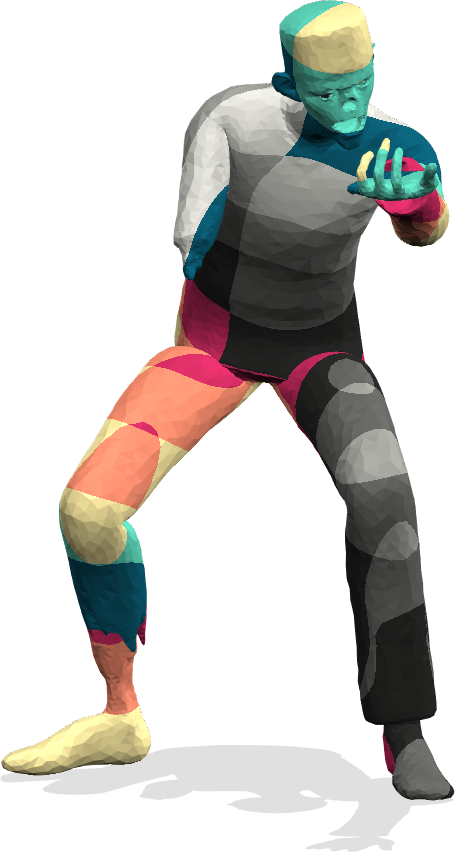}
\end{tabular}
    \caption{\textbf{Qualitative results of our method on DT4D-H inter-class.} Our method obtains accurate correspondences for non-isometrically deformed shapes.}
    \label{fig:dt4d}
\end{figure}

\end{document}